\documentclass[twocolumn,runningheads]{svjour3}

\usepackage{amssymb}
\usepackage{graphicx}

\usepackage{amsmath,amssymb,amscd} 
\def\bbbr{{\mathbb R}} 
\def\bbbz{{\mathbb Z}}

\usepackage{url}

\journalname{arXiv preprint}

\begin{document}

\title{Scale-covariant and scale-invariant Gaussian derivative networks%
\thanks{The support from the Swedish Research Council (contract 2018-03586) is gratefully acknowledged.}}

\titlerunning{Scale-covariant and scale-invariant Gaussian derivative networks}

\author{Tony Lindeberg}

\institute{Computational Brain Science Lab,
                Division of Computational Science and Technology,
                KTH Royal Institute of Technology,
                SE-100 44 Stockholm, Sweden.
                \email{tony@kth.se}}

              \date{}
              
\maketitle

\begin{abstract}
This paper presents a hybrid approach between scale-space theory and
deep learning, where a deep learning architecture is constructed by
coupling parameterized scale-space operations in cascade.
By sharing the learnt parameters between multiple scale channels, and
by using the transformation properties of the scale-space primitives
under scaling transformations, the resulting network becomes provably
scale covariant. By in addition performing max pooling over the
multiple scale channels, or other permutation-invariant pooling over
scales, a resulting network architecture for image
classification also becomes provably scale invariant.

We investigate the performance of such networks on the MNIST Large Scale
dataset, which contains rescaled images from the original MNIST
dataset over
a factor of 4 concerning training data and over a factor of 16 concerning
testing data. It is demonstrated that the resulting approach allows for scale
generalization, enabling good performance for classifying patterns at
scales not spanned by the training data.
%

\end{abstract}

\section{Introduction}

Variations in scale constitute a substantial source of variability in
real-world images, because of objects having different size in the
world and being at different distances to the camera.

A problem with traditional deep networks, however, is that they are not
covariant with respect to scaling transformations in the image domain.
In deep networks, non-linear\-ities are performed relative to the
current grid spacing, which implies that the deep network does not 
commute with scaling transformations.
Because of this lack of ability to handle scaling variations in the
image domain, the performance of deep networks may be very poor
when subject to testing data at scales that are not spanned by the
training data.

One way of achieving scale covariance in a brute force manner is by
applying the same deep net to multiple rescaled copies of the input
image. Such an approach is developed and investigated in
\cite{JanLin21-ICPR+arXiv}. When working with such a scale-channel
network it may, however, be harder to combine information between the
different scale channels, unless the multi-resolution representations at
different levels of resolution are also resampled to a common
reference frame when information from different scale levels 
is to be combined.

Another approach to achieve scale covariance is by applying multiple rescaled non-linear
filters to the same image. For such an architecture, it will
specifically be easier to combine information from multiple scale
levels, since the image data at all scales have the same resolution.

For the primitive discrete filters in a regular
deep network, it is, however, not obvious how to rescale the primitive
components, in terms of {\em e.g.\/}, local $3 \times 3$ or
$5 \times 5$ filters or max pooling over $2 \times 2$ neighbourhoods
in a sufficiently accurate manner over continuous variations
of spatial scaling factors. For this reason, it would be preferable to have a continuous
model of the image filters, which are then combined together
into suitable deep architectures, since
the support regions of the continuous filters could then be rescaled in a
continuous manner. Specifically, if we choose these filters as
scale-space filters, which are designed to handle scaling
transformations in the image domain, we have the potential of constructing a rich
family of hierarchical networks based on scale-space operations
that are provably scale covariant \cite{Lin20-JMIV}.

The subject of this article is to develop and experimentally
investigate one such hybrid
approach between scale-space theory and deep learning. The idea that we shall follow
is to define the layers in a deep architecture from scale-space
operations, and then use the closed-form transformation properties
of the scale-space primitives under scaling transformations to achieve provable scale covariance and
scale invariance of the resulting continuous deep network. Specifically, we will demonstrate that this will
give the deep network the ability to generalize to previously unseen
scales that are not spanned by the training data. This generalization
ability implies that training can be performed at some scale(s) and
testing at other scales, within some predefined scale range, with
maintained high performance over substantial scaling variations.

Technically, we will experimentally explore this idea for one specific
type of architecture, where the layers are parameterized linear
combinations of Gaussian derivatives up to order two. With such a
parameterization of the filters in the deep network, we also obtain a
compact parameterization of the degrees of freedom in the network,
with of the order of 16~000 or 38~000
parameters for the networks used in the experiments in this paper,
which may be advantageous in situations when only smaller
sets of training data are available. The overall
principle for obtaining scale covariance and scale invariance 
is, however, much more general and applies to much wider classes of
possible ways of defining layers from scale-space operations.

\subsection{Structure of this article}

This paper is structured as follows:
Section~\ref{sec-related-work} begins with an overview of related
work, with emphasis on approaches for handling scale variations
in classical computer vision and deep networks.
Section~\ref{sec-scale-gen} introduces the technical material with a conceptual
overview description of how the notions of scale covariance and scale
invariance enable scale generalization, {\em i.e.\/}, the ability to
perform testing at scales not spanned by the training data.
Section~\ref{sec-gaussdernets} defines the notion of Gaussian derivative
networks, gives their conceptual motivation and proves their 
scale covariance and scale invariance properties.
Section~\ref{sec-exp-mnist} describes the result of applying a
single-scale-channel Gaussian derivative network to the regular MNIST
dataset.
Section~\ref{sec-exp-mnist-large-scale} describes the result of a
applying multi-scale-channel Gaussian derivative networks to the
MNIST Large Scale dataset, with emphasis on scale generalization
properties and scale selection properties.
Finally, Section~\ref{sec-summ-disc} concludes with a summary and discussion.

\subsection{Relations to previous contribution}

This paper is an extended version of a paper presented at the SSVM
2021 conference \cite{Lin21-SSVM} and with substantial additions
concerning:
\begin{itemize}
\item
  a wider overview of related work (Section~\ref{sec-related-work}),
\item
  a conceptual explanation about the potential advantages of scale
  covariance and scale invariance for deep networks, specifically with
  regard to how
  these notions enable scale generalization
  (Section~\ref{sec-scale-gen}),
\item
  more detailed mathematical definitions regarding the foundations of
  Gaussian derivative networks (Section~\ref{sec-def-gauss-der-nets})
  as well as more detailed proofs regarding their scale covariance
  (Section~\ref{sec-sc-cov}) and scale invariance
  (Section~\ref{sec-sc-inv}) properties,
\item
  a more detailed treatment of the scale selection properties of the
  resulting scale channel networks (Section~\ref{sec-sc-sel})
    well as a discussion about issues to consider when training
  multi-scale-channel networks (Section~\ref{sec-mult-sc-train}).
\end{itemize}
In relation to the SSVM 2021 paper, this paper therefore (i)~gives a more
general treatment about the importance of scale generalization,
(ii)~gives a more detailed treatment about the theory of the presented
Gaussian derivative networks that could not be included in the
conference paper because of the space limitations, (iii)~describes scale
selection properties of the resulting scale channel networks and
(iv)~gives overall better descriptions of the subjects treated in the paper,
including (v)~more extensive references to related literature.

\section{Relations to previous work}
\label{sec-related-work}

In classical computer vision, it has been demonstrated that
scale-space theory constitutes a powerful paradigm for constructing
scale-covariant and scale-invariant feature detectors and making
visual operations robust to scaling transformations
\cite{Lin97-IJCV,Lin98-IJCV,BL97-CVIU,ChoVerHalCro00-ECCV,MikSch04-IJCV,Low04-IJCV,BayEssTuyGoo08-CVIU,TuyMik08-Book,Lin13-ImPhys,Lin15-JMIV}.
In the area of deep learning, a corresponding framework for handling
general scaling transformations has so far not been as well established.

Concerning the relationship between deep networks and scale, several researchers
have observed robustness problems of deep networks under scaling
variations \cite{FawFro15-BMVC,SinDav18-CVPR}.
There have been some approaches developed to aim at 
scale-invariant convolutional neural networks (CNNs)
\cite{XuXiaZhaYanZha14-arXiv,KanShaJac14-arXiv,MarKelLobTui18-arXiv,GhoGup19-arXiv}.
These approaches have, however, not been experimentally
evaluated on the task of generalizing to
scales not present in the training data 
\cite{KanShaJac14-arXiv,MarKelLobTui18-arXiv}, or only over a very
narrow scale range \cite{XuXiaZhaYanZha14-arXiv,GhoGup19-arXiv}.

For studying scale invariance in a more general setting, we argue that
it is essential to experimentally verify the scale invariant
properties over sufficiently wide scale ranges. For scaling factors
moderately near one, a network could in principle learn to handle scaling
transformations by mere training, {\em e.g.\/}, by data augmentation of
the training data. For wider ranges of scaling factors, that
will, however, either not be possible or at least very inefficient, if
the same network, with a uniform internal architecture, is to
represent both a large number of very fine-scale and a large number of
very coarse-scale image structures.
Currently, however, there is a lack of datasets that both cover 
sufficiently wide scale ranges and contain sufficient amounts of data
for training deep networks. For this reason, we have in a companion
work created the MNIST Large Scale dataset
\cite{JanLin21-ICPR,JanLin20-MNISTLargeScale}, which covers scaling
factors over a range of 8, and over which we will evaluate the
proposed methodology, compared to previously reported experimental
work that cover a scale range of
the order of a factor of 3
\cite{XuXiaZhaYanZha14-arXiv,GhoGup19-arXiv,SosSzmSme20-ICLR}.

Provably scale-covariant or scale-equivariant%
\footnote{In the deep learning literature, the terminology ``scale
  equivariant'' has become common for what is called ``scale
  covariance'' in scale-space theory. In this paper, we use the
  terminology ``scale covariance'' to keep consistency with the
  earlier scale-space literature \cite{Lin13-ImPhys}.} networks that incorporate the
transformation properties of the network under scaling
transformations, or approximations thereof,
have been recently developed in
\cite{WorWel19-NeuroIPS,Lin19-SSVM,Lin20-JMIV,SosSzmSme20-ICLR,Bek20-ICLR,JanLin21-ICPR+arXiv,Lin21-SSVM}.
In principle, there are two types of approaches to achieve scale
covariance by expanding the image data over the scale dimension:
(i)~either by applying multiple rescaled filters to each input image or
(ii)~by rescaling each underlying image
over multiple scaling factors and applying the same deep network to
all these rescaled images.
In the continuous case, these two dual approaches are computationally
equivalent, whereas they may differ in practice depending upon how the
discretization is done and depending upon how the computational components
are integrated into a composed network architecture.

Still, all the components of a scale-covariant or scale-invariant
network, also the training stage
and the handling of boundary effects in the scale direction, need to
support true scale invariance or a sufficiently good approximation thereof,
and need to be experimentally verified
on scale generalization tasks, in order to support true scale
generalization.%
\footnote{For example, in a companion work
  \cite{JanLin21-ICPR+arXiv}, we noticed that for an alternative sliding
  window approach studied in
   that work, full scale generalization is not achieved, because
   the support regions of the receptive fields in the training stage
   do not handle all the sizes of the input in a uniform manner, thereby
   negatively affecting the scale generalization properties, although
   the overall network architecture would otherwise support true scale invariance.}

A main purpose of this article is to demonstrate how provably scale-covariant and
scale-invariant deep networks can be constructed using straightforward
extensions of established
concepts in classical scale-space theory,
and how the resulting
networks enable very good scale generalization. Extensions of ideas and
computational mechanisms in classical scale-space theory to handling
scaling variations in deep networks for object recognition have also
been recently explored in \cite{SinNajShaDav21-PAMI}. Using a set of
parallel scale channels, with shared weights between the scale
channels, as used in other scale-covariant networks as well as in this work,
has also been recently explored
for object recognition in \cite{LiCheWanZha19-ICCV}.

In contrast to some other work, such as \cite{WorWel19-NeuroIPS}, the
scale-covariant and scale-invariant properties of our networks are not restricted to
scaling transformations that correspond to integer scaling
factors. Instead, true scale covariance and scale invariance hold for
the scaling factors that correspond to ratios of the scale values of
the associated scale channels in the network.
For values in between, the results will
instead be approximations, whose accuracy will also be determined by
the network architecture and the training method.

In the implementation underlying this
work, we sample the space of scale channels by multiples of
$\sqrt{2}$, which has also been previously demonstrated to be a good
choice for many classical scale-space algorithms, as a trade-off
between computational accuracy, as improved by decreasing this ratio,
and computational efficiency, as improved by increasing this ratio.
In complementary companion work for the dual approach of constructing
scale covariant deep networks by expansion over rescaled input images for multiple
scaling factors \cite{JanLin21-ICPR+arXiv}, this choice of sampling
density has also been shown to be a good trade-off in the sense that
the accuracy of the network is improved substantially by decreasing
the scale sampling factor from a factor of $2$ to $\sqrt{2}$, however,
very marginally by decreasing the scale sampling factor further again to
$\sqrt[4]{2}$.

A conceptual simplification that we shall make, for simplicity of
implementation and experimentation only, is that all the filters
between the layers in the network operate only on information within
the same scale channel. A natural generalization to consider, would be
to also allow the filters to also access information from neighbouring
scale channels%
\footnote{With regard to the Gaussian derivative networks described in
  this article, the transformation between adjacent layers in
  Equation~(7), should then be expanded with a sum of corresponding
  contributions from a set of neighbouring scale channels, in order to
allow for interactions between image information at different scales.},
as used in several classical scale-space methods
\cite{SchCro00-IJCV,LinLin04-ICPR,LapLin04-ECCVWS,LinLin12-CVIU,LarDarDahPed12-ECCV}
and also used within the notion of group convolution in the deep learning
literature \cite{CohWel16-ICML}, which would then enable interactions between 
image information at different scales. A possible problem with allowing the
filters to extend to neighbouring scale channels, however, is that the
boundary effects in the scale direction,
caused by a limited number of scale channels,
may then become more problematic, which may affect the scale generalization
properties. For this reason, we restrict ourselves to
within-scale-channel processing only, in our first implementation, although the conceptual
formulation would with straightforward extensions also apply to scale
interactions and group convolutions.

Spatial transformer networks have been proposed as a general approach
for handling image transformations in CNNs
\cite{JadSimZisKav15-NIPS,LinLuc17-CVPR}. 
Plain transformation of the feature maps in a CNN by a spatial transformer will,
however, in general, not 
deliver a correct transformation and will therefore not support truly invariant
recognition, although
spatial transformer networks that instead operate by transforming the
input will allow for true transformation-invariant properties, at the
cost of more elaborate network architectures with associated possible
training problems \cite{FinJanLin21-ICPR+arXiv,JanMayFinLin20-arXiv}.

Concerning other deep learning approaches that somehow handle or are related to the notion of scale, 
deep networks have been applied to the multiple layers in an image pyramid
\cite{SerEigZhaMatFerLeC13-arXiv,Gir15-ICCV,LinDolGirHeHarBel17-CVPR,LinGoyGirHeDol17-ICCV,HeKiDolGir17-ICCV,HuRam17-CVPR}, 
or using other multi-channel approaches where the input image is rescaled to different
resolutions, possibly combined with
interactions or pooling between the layers \cite{RenHeGirZhaSun16-PAMI,NahKimLee17-CVPR,ChePapKokMurYui17-PAMI}.
Variations or extensions of this approach include scale-dependent
pooling \cite{YanChoLin16-CVPR}, using sets of
subnetworks in a multi-scale fashion \cite{CaiFanFerVas16-ECCV},
dilated convolutions
\cite{YuKol16-ICLR,YuKolFun17-CVPR,MehRasCasShaHaj18-ECCV},
scale-adaptive convolutions \cite{ZhaTanZhaLiYan17-ICCV} or
adding additional branches of down-samplings and/or up-samplings in each
layer of the network
\cite{WanKemFarYuiRas19-CVPR,CheFanXuYanKalRohYanFen19-ICCV}.
The aims of these types of networks have, however, not been to achieve
scale invariance in a strict mathematical sense, and they could rather
be coarsely seen as different types of approaches to
enable multi-scale processing of the image data in deep networks.

In classical computer vision, receptive field models in terms of
Gaussian derivatives have been demonstrated to constitute a powerful
approach to handle image structures at multiple scales, where the
choice of Gaussian derivatives as primitives in the first layer of
visual processing can be motivated by mathematical
necessity results \cite{Iij62,Wit83,Koe84,BWBD86-PAMI,KoeDoo92-PAMI,Lin93-Dis,Lin94-SI,Flo97-book,WeiIshImi99-JMIV,Haa04-book,Lin10-JMIV,Lin13-BICY}.

In deep networks, mathematically defined models of receptive fields
have been used in different ways.
Scattering networks have been proposed based on Morlet wavelets
\cite{BruMal13-PAMI,SifMal13-CVPR,OyaMal15-CVPR}.
Gaussian derivative kernels have been used as structured receptive
fields in CNNs \cite{JacGemLouSme16-CVPR}.
Gabor functions have been proposed as primitives for
modulating learned filters \cite{LuaCheZhaHanLiu18-PAMI}.
Affine Gaussian kernels have been used to
compose free-form filters to adapt the receptive field size and shape to the image data \cite{SheWanDar19-arXiv}.

As an alternative approach to handle scaling transformations in deep networks, the image data has
been spatially warped by a log-polar transformation prior to the image
filtering steps
\cite{HenVed17-ICML,EstAllZhoDan18-ICLR}, implying that the scaling
transformation is mapped to a mere translation in the
log-polar domain. Such a log-polar transformation does, however,
violate translational covariance over the original spatial domain,
as otherwise obeyed by a regular CNN applied to the original input image.

Approaches to handling image transformations in deep networks have
also been developed based on formalism from group theory
\cite{PogAns16-book,LapSavBuhPol16-CVPR,CohWel16-ICML,KonTri18-ICML}.
A general framework for handling basic types of natural image
transformations in terms of spatial scaling transformation, spatial affine
transformations, Galilean transformations and temporal scaling
transformations  in the first layer of visual processing based on
linear receptive fields and with relations to biological vision has been presented in \cite{Lin21-Heliyon}, based on generalized axiomatic
scale-space theory \cite{Lin13-ImPhys}.

The idea of modelling layers in neural networks as continuous
functions instead of discrete filters has also been advocated in
\cite{LeRBen07-AISTATS,WanSuoMaPokUrt18-CVPR,WuQiFux19-CVPR,ShoFeiHaiIra20-arXiv}.
The idea of reducing the number of parameters of deep networks by a compact
parameterization of continuous filter shapes has also been recently
explored in \cite{DuiSmeBekPor21-SSVM}, where PDE layers are defined as
parameterized combinations of diffusion, morphological and transport
processes and are demonstrated to lead to a very compact
parameterization. Conceptually, there are structural similarities between
such PDE-based networks and the Gaussian derivative networks that we
study in this work in the
sense that: (i)~the Gaussian smoothing underlying the Gaussian
derivatives correspond to a diffusion process and (ii)~the ReLU
operations between adjacent layers correspond to a special case of the
morphological operations, whereas the approaches differ in the sense
that (iii)~the Gaussian derivative networks do not contain an explicit
transport mechanism, while the primitive kernels in the Gaussian
derivative layers instead comprise explicit spatial oscillations or local ripples.
Other types of continuous models for deep networks in terms of PDE
layers have been studied in \cite{RutHab20-JMIV,SheHeLinMa20-ICML}.

Concerning the notion of scale covariance and its relation to scale generalization, a general sufficiency
result was presented in \cite{Lin20-JMIV} that guarantees provable
scale covariance for hierarchical networks that are constructed from
continuous layers defined from partial derivatives or differential
invariants expressed in terms of scale-normalized derivatives.
This idea was developed in more detail for a hand-crafted quasi quadrature
network, with the layers representing idealized models of complex
cells, and experimentally applied to the task of texture classification.
It was demonstrated that the resulting approach allowed for scale
generalization on the KTH-TIPS2 dataset, enabling classification of
texture samples at scales not present in the training data.

Concerning scale generalization for CNNs, \cite{JanLin21-ICPR}
presented a multi-scale-channel approach, where the same discrete CNN was applied
to multiple rescaled copies of each input image. It was demonstrated
that the resulting scale-channel architectures had much better ability to
handle scaling transformations in the input data compared to a regular vanilla CNN, and also that
the resulting approach lead to good scale generalization, for
classifying image patterns at scales not spanned by the training data.

The subject of this article is to complement the latter works, and
specifically combine a specific instance of the general class of
scale-covariant networks in \cite{Lin20-JMIV}  with deep learning and
scale channel networks \cite{JanLin21-ICPR}, where
we will choose the continuous layers as linear combinations of
Gaussian derivatives and demonstrate how such an architecture allows
for scale generalization.

\begin{figure*}
 \begin{center}
    \begin{tabular}{cc}
      {\small\em Fixed receptive field size\/} & {\small\em Matching receptive field sizes\/} \\
       \includegraphics[width=0.45\textwidth]{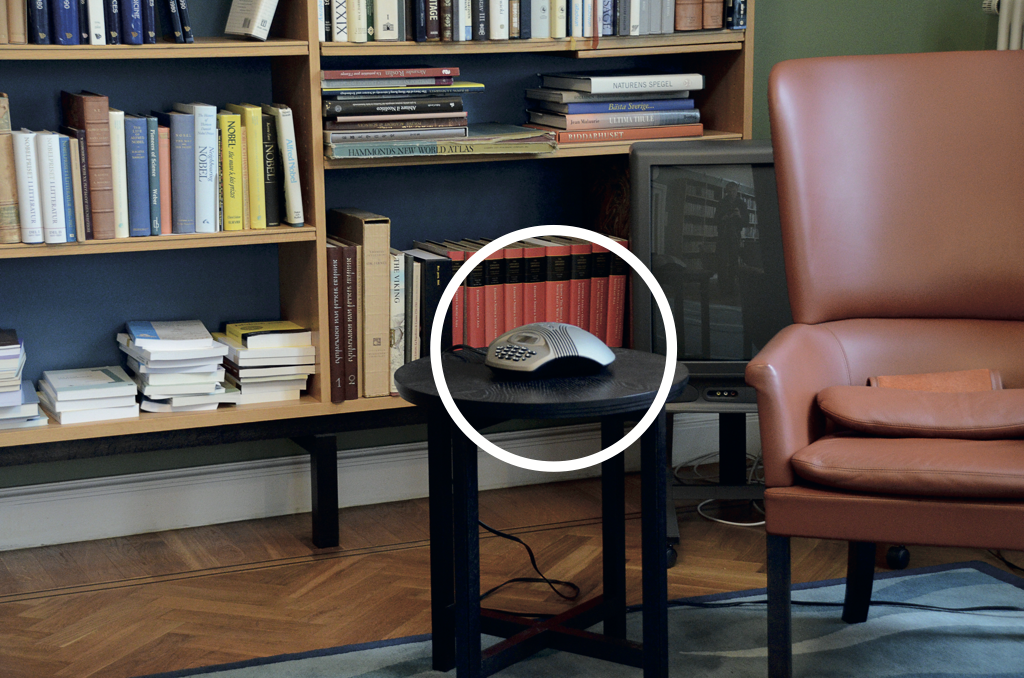}
       & \includegraphics[width=0.45\textwidth]{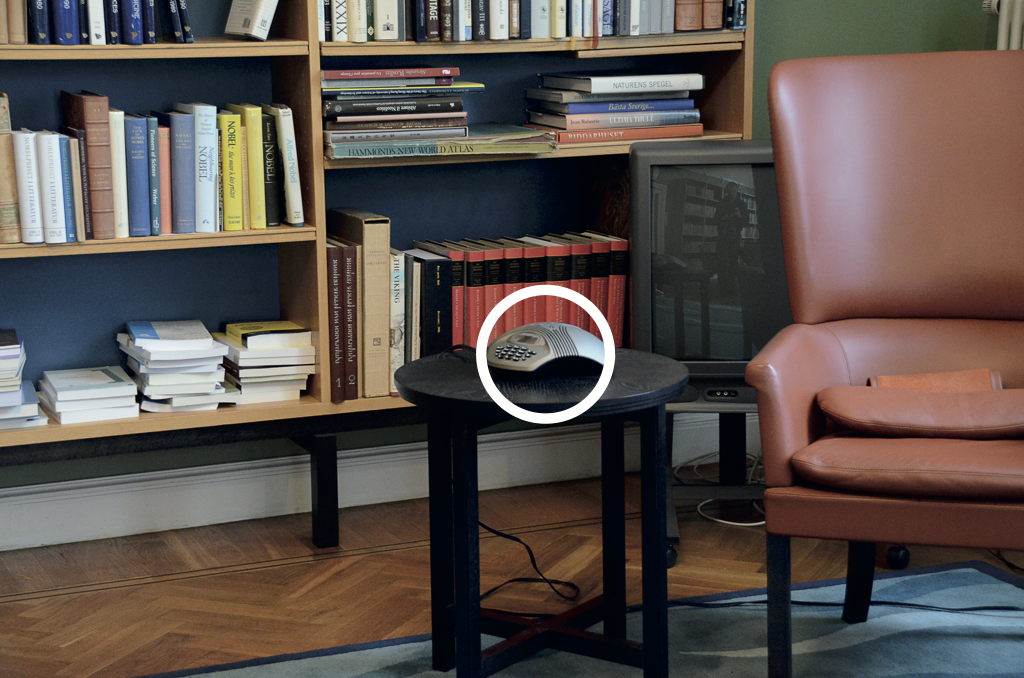} \\
       \includegraphics[width=0.45\textwidth]{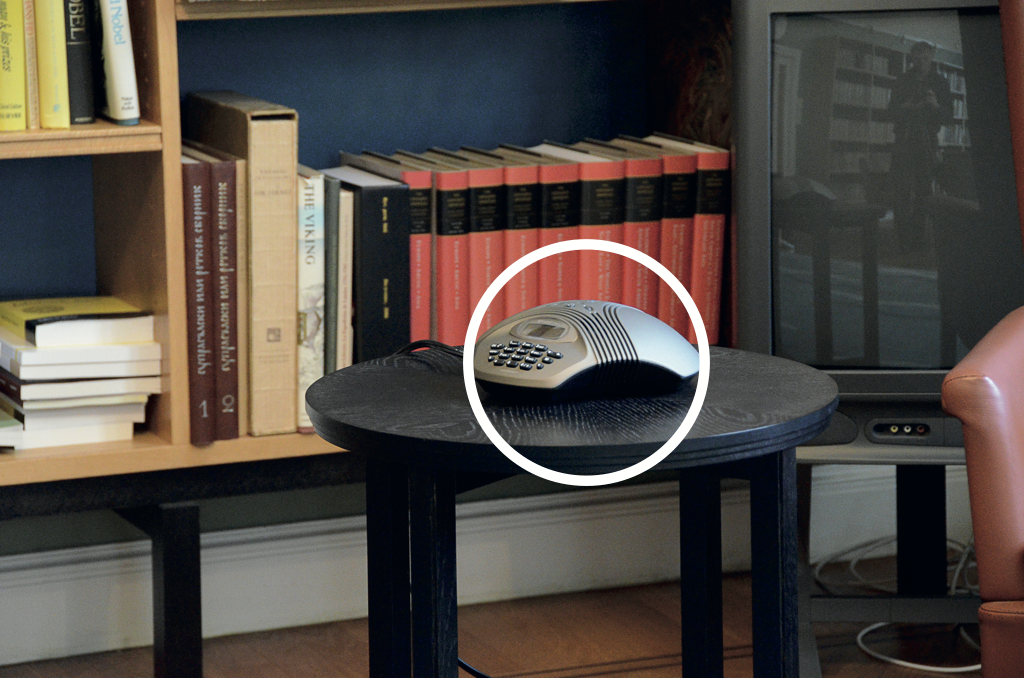}
       & \includegraphics[width=0.45\textwidth]{zoom2.png} \\
           \includegraphics[width=0.45\textwidth]{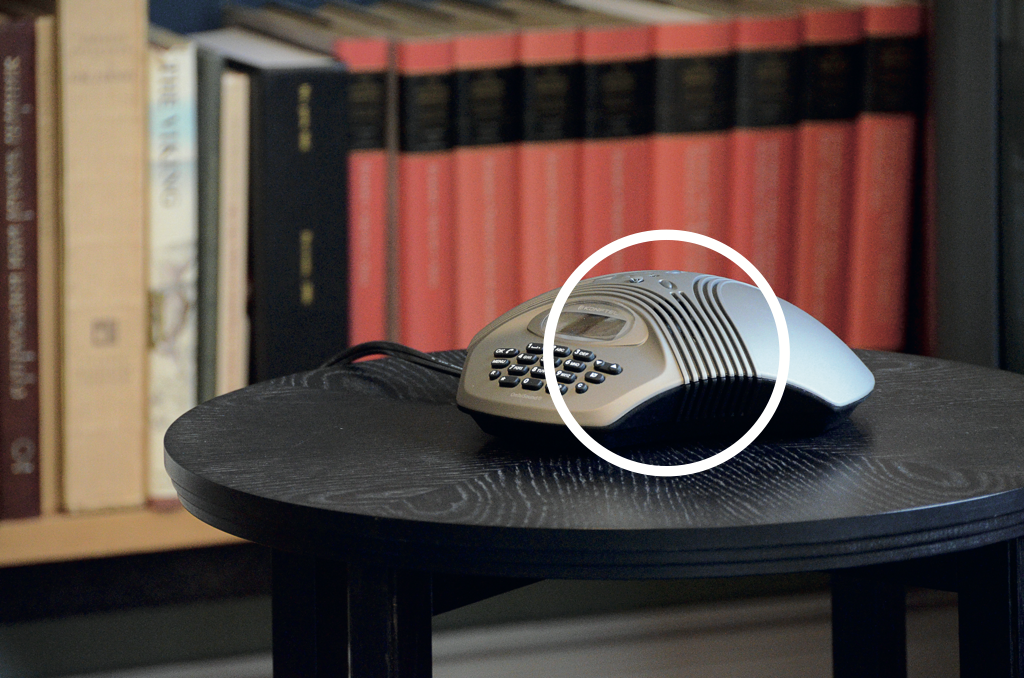}
       & \includegraphics[width=0.45\textwidth]{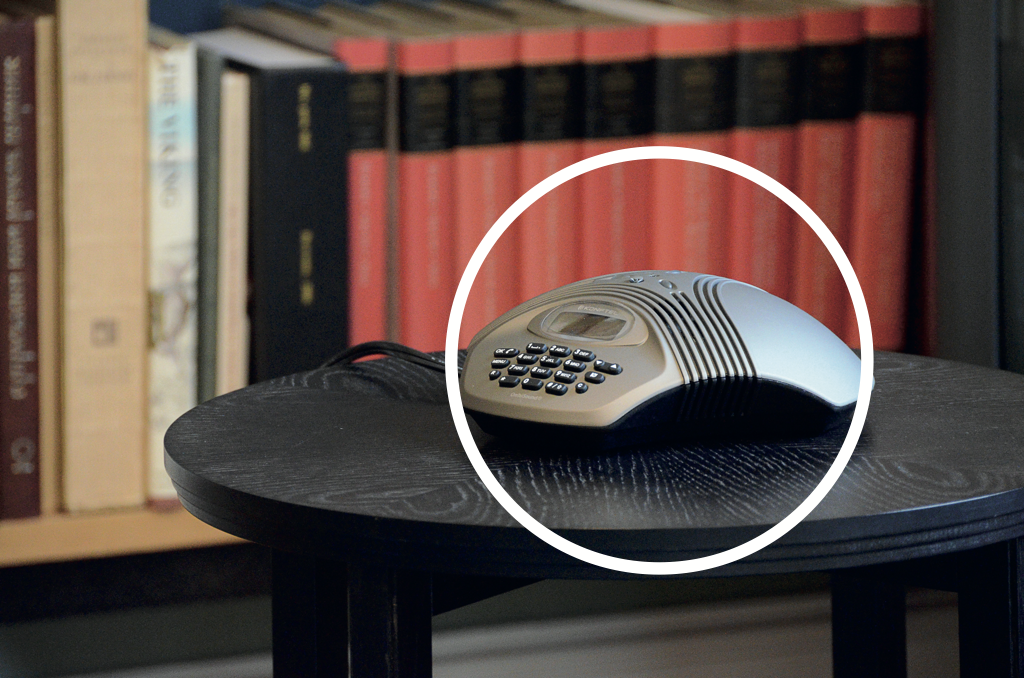} \\
        \end{tabular}
  \end{center}
  \caption{Illustration of the importance of having matching support
    regions of receptive fields when handling scaling transformations
    in the image domain. For this figure, we have simulated the effect
    of varying the distance between the object and the camera by
    varying the amount of zoom for a zoom lens.
    The left column illustrates the effect of having a fixed receptive
  field size in the image domain, and how that fixed receptive field
  size affects the backprojected receptive fields in the world, if
  there are significant scale variations.
  In the right column, the receptive field sizes are matched under the
scaling transformation, as enabled by scale covariant receptive field
families and scale channel networks, which makes it possible to define
deep networks that are invariant to scaling transformations, which in turn
enables scale generalization.}
  \label{fig-support-regions-zoom}
\end{figure*}

\section{Scale generalization based on scale covariant and scale
  invariant receptive fields}
\label{sec-scale-gen}

A conceptual problem that we are interested in addressing is to be
able to
perform {\em scale generalization\/}, {\em i.e.\/}, being able to
train a deep network for image structures at some scale(s), and then
being able to test the network at other scales, with maintained high
recognition accuracy, and without complementary
use of data augmentation.

The motivation for addressing this problem is that scale variations
are very common in real-world image data, because of objects being of
different sizes in the world, and because of variations in the viewing
distance to the camera. Regular CNNs are, however, known to perform
very poorly when exposed to testing data at scales that are not
spanned by the training data.

Thus, a main goal of this work is to equip deep networks with prior
knowledge to handle scaling transformations in image data, and
specifically the ability to generalize to {\em new scales that are not
  spanned by the training data\/}.

The methodology that we will follow to achieve such scale generalization
is to require that the image descriptors or the receptive fields in the
deep network are to be truly scale covariant and scale invariant.
The underlying idea is that the output from the image processing
operations in the deep network should thus remain sufficiently similar
under scaling transformations, as will be later formalized into the
commutative diagram in Figure~\ref{fig-comm-diag-sc-cov}.
Operationalized, we will require that the receptive field responses
can be matched under scaling transformations.
Specifically, we may require that the support regions of the receptive
fields can be matched under scaling transformations.

Figure~\ref{fig-support-regions-zoom} illustrates these effects for
scaling transformations in an indoor scene, caused by varying the
amount of zoom for a zoom lens, which leads to similar scaling
transformations as when varying the distance between the object and
the camera.
In the left column, we have marked receptive fields of constant size.
Then, because of the scaling transformations, the backprojections of
these receptive fields will be different in the world because of the
scaling transformation, which may cause problems for a deep network,
depending on how it is constructed. In the right column, the support
regions follow the scaling transformations in a matched and scale covariant way,
making it much easier for a deep network to recognize the object of
interest under scaling transformations.

\subsection{Scale covariance and scale invariance}

With the scaling operator%
\footnote{From the view-point of group-equivariant CNNs
  \cite{CohWel16-ICML} as applied to constructing scale-covariant or scale-equivariant
  CNNs based on formalism from group theory
  \cite{WorWel19-NeuroIPS,SosSzmSme20-ICLR,Bek20-ICLR}, this scaling
  operator can be seen as corresponding to a representation of the
  scaling group. In the more technical presentation that follows later
  in Section~\ref{sec-gaussdernets},
  the influence of this scaling group will be in terms of an analysis
  of the transformation properties under scaling transformations of
  the receptive field responses in the studied class of deep networks.}
${\cal S}_s$ defined by
\begin{equation}
    ({\cal S}_s f)(x) = f(S_s^{-1} x) = f(\tfrac{x}{s}).
\end{equation}
and with $\Gamma$ denoting a family of feature map operators, 
the notion of {\em scale covariance\/}
means that the feature maps should commute with scaling
transformations and transform
according to
\begin{equation}
   \Gamma'({\cal S}_s f) = {\cal S}_s (\Gamma(f)),
\end{equation}
where $\Gamma'$ represents some possibly transformed feature map
operator within the same family $\Gamma$, and adapted to the rescaled image domain.
If we parameterize the feature map operator $\Gamma$ by a scale
parameter $s$, we can make a commutative diagram as shown in
Figure~\ref{fig-comm-diag-sc-cov}.

\begin{figure}
  \begin{center}
  \[
    \begin{CD}
       (\Gamma_s f)(x)
       @>{\footnotesize 
              s' = S s, \, \Gamma_s(f)(x)=\Gamma_{s'}(f')(x')}>> (\Gamma_{s'} f')(x') \\
       \Big\uparrow\vcenter{\rlap{$\scriptstyle{{\Gamma_s}}$}} & &
       \Big\uparrow\vcenter{\rlap{$\scriptstyle{{\Gamma_{s'}}}$}} \\
       f(x) @>{\footnotesize \footnotesize  x' = S x, \, f'(x') = f(x) }>> f'(x')
    \end{CD}
  \]
  \end{center}
  \caption{Commutative diagram for a scale-parameterized feature map
    operator $\Gamma_s$ that is applied to image data under scaling
    transformations. The commutative diagram, which should be read
    from the lower left corner to the upper right corner, means that
    irrespective of whether the input image is first subject to a
    scaling transformation and then the computation of a feature map,
    or whether the feature map is computed first and then transformed
    by a scaling transformation, we should get the same result.
    Note, however, that this definition of scale covariance assumes a
    multi-scale representation of the image data, and that direct
    availability to the image
    representations at the matching scale levels $s' = S s$ is necessary to complete
  the commutative diagram.}
  \label{fig-comm-diag-sc-cov}
\end{figure}


Implicit in Figure~\ref{fig-comm-diag-sc-cov} is the notion that the deep network performs
processing at multiple scales.
The motivation for such a multi-scale approach is that for a vision system
that observes an {\em a priori\/} unknown scene, there is no way to
know in advance what scale is appropriate for processing the image data.
In the absense of further information about what scales are
appropriate, the only reasonable approach is to consider image
representations at multiple scales. This motivation is similar to a
corresponding motivation underlying scale-space theory 
\cite{Iij62,Wit83,Koe84,BWBD86-PAMI,KoeDoo92-PAMI,Lin93-Dis,Lin94-SI,Flo97-book,WeiIshImi99-JMIV,Haa04-book,DuiFloGraRom04-JMIV,Lin10-JMIV,Lin13-ImPhys},
which has developed a systematic approach for handling scaling
transformations and image structures at multiple scales for
hand-crafted image operations.
Regarding deep networks that are to learn their image representations
from image data, we proceed in a conceptually similar manner,
by considering deep networks with {\em multiple scale channels\/},
that perform similar types of processing operations in all the scale
channels, although over multiple scales.

Scale invariance does in turn mean that the final output from the deep
network $\Lambda$, for example the result of max pooling or average
pooling over the multiple scale channels, should not in any way be be
affected by scaling transformations in the input
\begin{equation}
    \Lambda({\cal S}_s f) = \Lambda(f),
\end{equation}
at least over some predefined range of scale that defines the capacity
of the system,
where scale covariance of the receptive fields in the lower layers in the
network makes scale invariance possible in the final layer(s).

\subsection{Approach to scale generalization}

By basing the deep network on image operations that are
provably scale covariant and scale invariant, we can train on some scale(s) and
test at other scale(s). 

The scale covariant and scale invariant properties of the image operations will then 
allow for {\em transfer\/} between image information at
different scales.

In the following, we will develop this approach for one specific class
of deep networks.
Conceptually similar approaches to scale generalization of deep
networks are applied in
\cite[Figures~15--16]{Lin20-JMIV} and \cite{JanLin21-ICPR}.
Conceptually similar approaches to scale invariance and scale generalization for
hand-crafted computer vision operations are applied in
\cite{Lin97-IJCV,Lin98-IJCV,BL97-CVIU,ChoVerHalCro00-ECCV,MikSch04-IJCV,Low04-IJCV,BayEssTuyGoo08-CVIU,TuyMik08-Book,Lin15-JMIV,Lin13-PONE}.

\subsection{Influence of the inner and the outer scales of the image}

Before starting with the technical treatment, let us, however, remark
that when designing scale-covariant and scale-invariant image
processing operations, for a given discrete image there will be only a
finite range of scales over which sufficiently good approximations to
continuous scale covariance and scale invariant properties can be
expected hold.

For a discrete image of finite resolution, there may at finer scales
be interference with the inner scale of the image, given by the
resolution of the sampling grid. For example, for the set of gradually
zoomed images in Figure~\ref{fig-support-regions-zoom}, the surface
texture is clearly visible to the observer in the most zoomed-in
image, whereas the same surface structure is far less visible to the observer in the most
zoomed-out image. Correspondingly, if we would zoom in further to the
telephone in the center of the image, peripheral parts of it may fall
outside the image domain, corresponding to interference with the outer
scale of the image, given by the image size. Sufficiently good numerical approximations to
scale covariance and scale invariance can therefore only be expected
to hold over a subrange of the
scale interval for which the relevant scales of the interesting image
structures are sufficiently within the range of scales defined by the
inner and the outer scale of the image.

In the experiments to be presented in this work, we will handle these
issues by a manual choice of the scale range for scale-space analysis
as determined by properties of the dataset.
When designing a computer vision system to analyze complex scenes with
a priori unknown scales in the image data, there may, however, in
addition be useful to add complementary mechanisms to handle the
fact that an object seen from a far distance may contain far less
fine-scale image structures than the same object seen from a nearby
distance by the same camera. This implies that the recognition
mechanisms in the system may need the ability to handle both the
presence and the absense of image information over different subranges
of scale, which goes beyond the continuous notions of scale covariance and
scale invariance studied in this work. In a similar way, some mechanism may be
needed to handle problems caused by the full object not being visible
within the image domain or with a sufficient margin around its
boundaries to reduce boundary effects of the spatially extended scale-space filters.
For the purpose of the presentation in
this paper, we will, however, leave automated handling of these
issues for future work.

\section{Gaussian derivative networks}
\label{sec-gaussdernets}

In a traditional deep network, the filter weights are usually free
variables with few additional constraints.
In scale-space theory, on the other hand, theoretical results have
been presented showing that Gaussian kernels and their corresponding
Gaussian derivatives constitute a canonical class of image operations%
\footnote{In this paper, we follow the school of scale-space
  axiomatics based on causality \cite{Koe84} or non-enhancement of
  local extrema \cite{Lin96-ScSp,Lin10-JMIV}, by which smoothing with
  the Gaussian kernel is the unique operation for scale-space
  smoothing, and Gaussian derivatives constitute a unique family
  of derived receptive fields \cite{KoeDoo92-PAMI,Lin93-Dis}. For the
  alternative school of scale-space axiomatics based on scale
  invariance \cite{Iij62}, a wider
  class of primitive smoothing operations is permitted
  \cite{PauFidMooGoo95-PAMI,FelSom04-JMIV,DuiFloGraRom04-JMIV},
  corresponding to the $\alpha$ scale spaces.}
\cite{Iij62,Wit83,Koe84,BWBD86-PAMI,KoeDoo92-PAMI,Lin93-Dis,Lin94-SI,Flo97-book,WeiIshImi99-JMIV,Haa04-book,Lin10-JMIV}.
In classical computer vision based on hand-crafted image features,
it has been demonstrated that a large number of visual tasks can be
successfully addressed by computing image features and image descriptors based on
Gaussian derivatives, or approximations thereof, as the first layer of image features
\cite{Lin97-IJCV,Lin98-IJCV,SchCro00-IJCV,MikSch04-IJCV,Low04-IJCV,BayEssTuyGoo08-CVIU,Lin15-JMIV}.
One could therefore raise the question if such Gaussian derivatives
could also be used as computational primitives for constructing deep
networks.

\begin{figure}[h]
  \begin{center}
    \begin{tabular}{c}
      \includegraphics[width=0.14\textwidth]{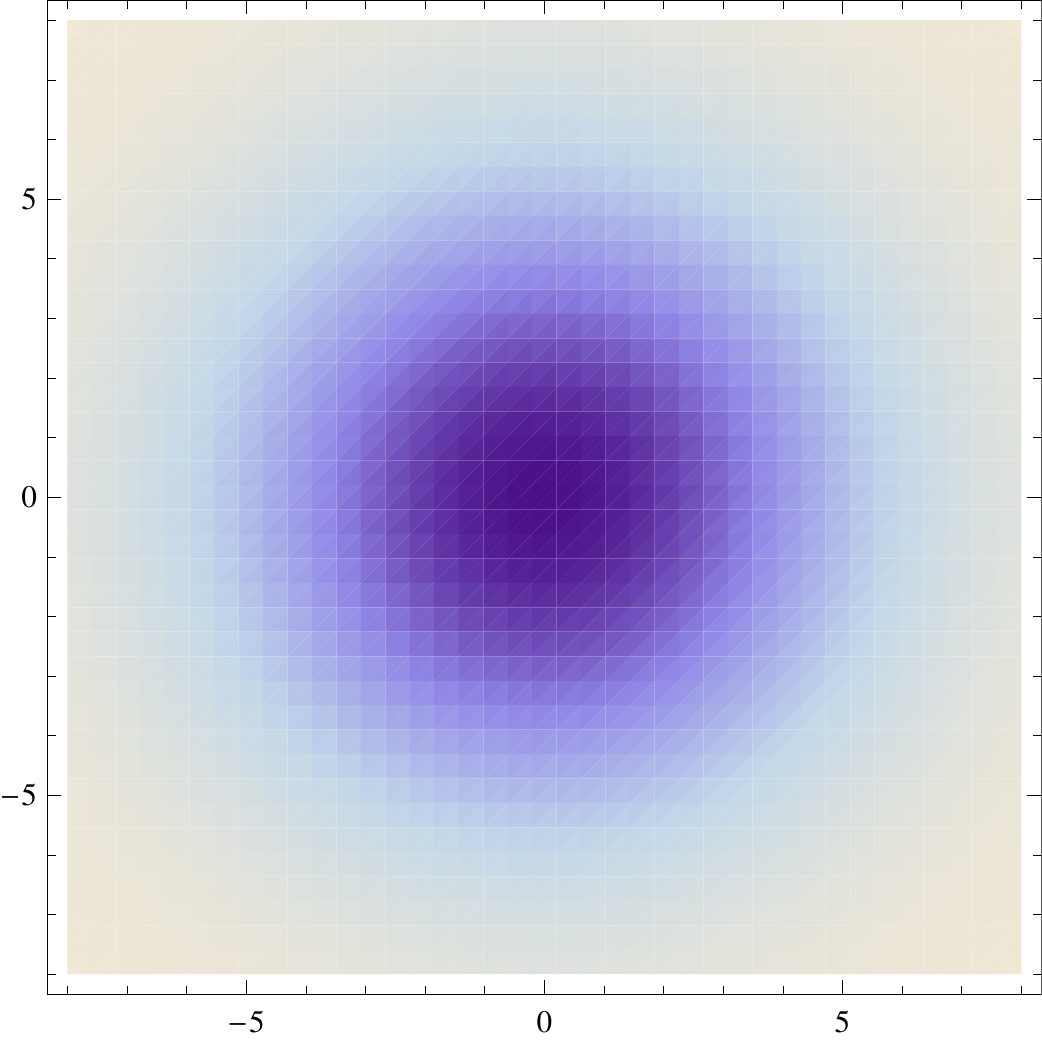} \\
    \end{tabular} 
  \end{center}
  \begin{center}
    \begin{tabular}{cc}
      \includegraphics[width=0.14\textwidth]{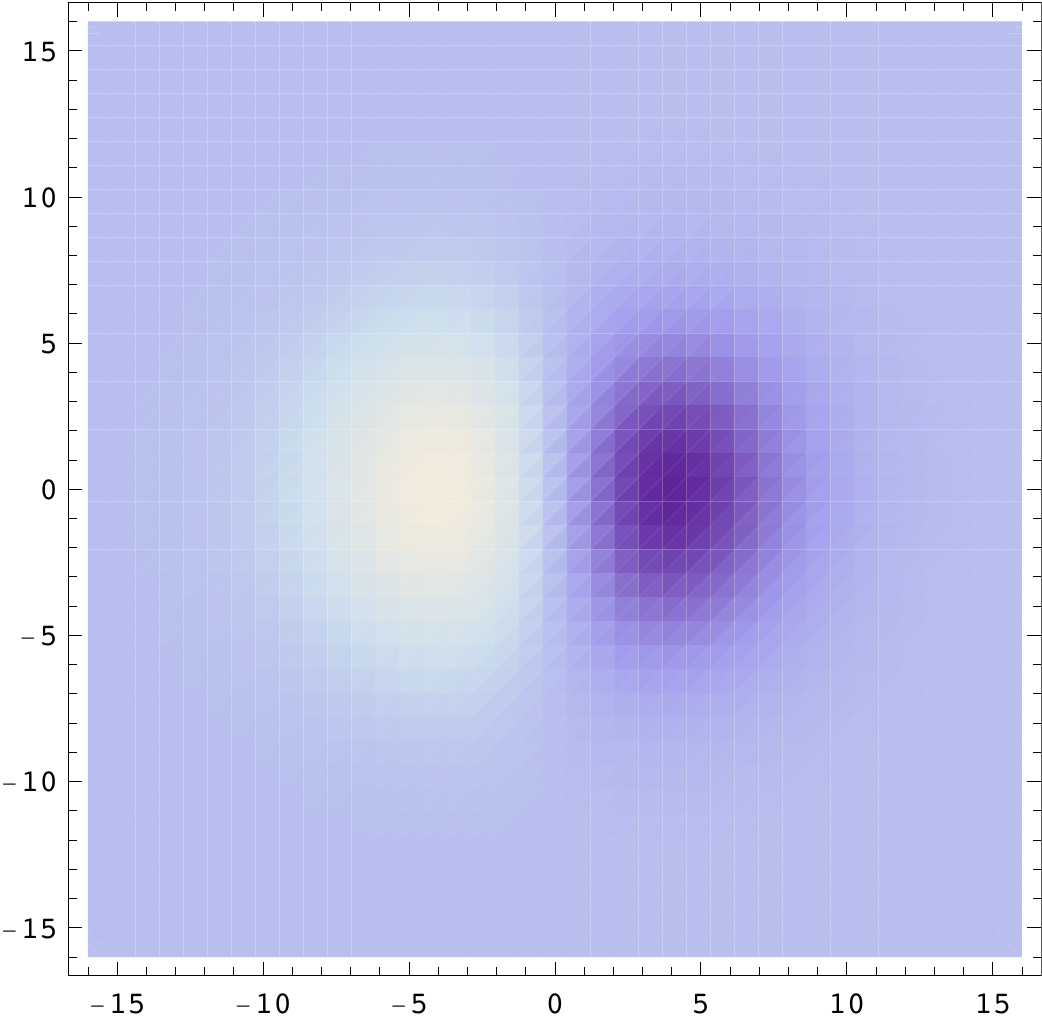} &
      \includegraphics[width=0.14\textwidth]{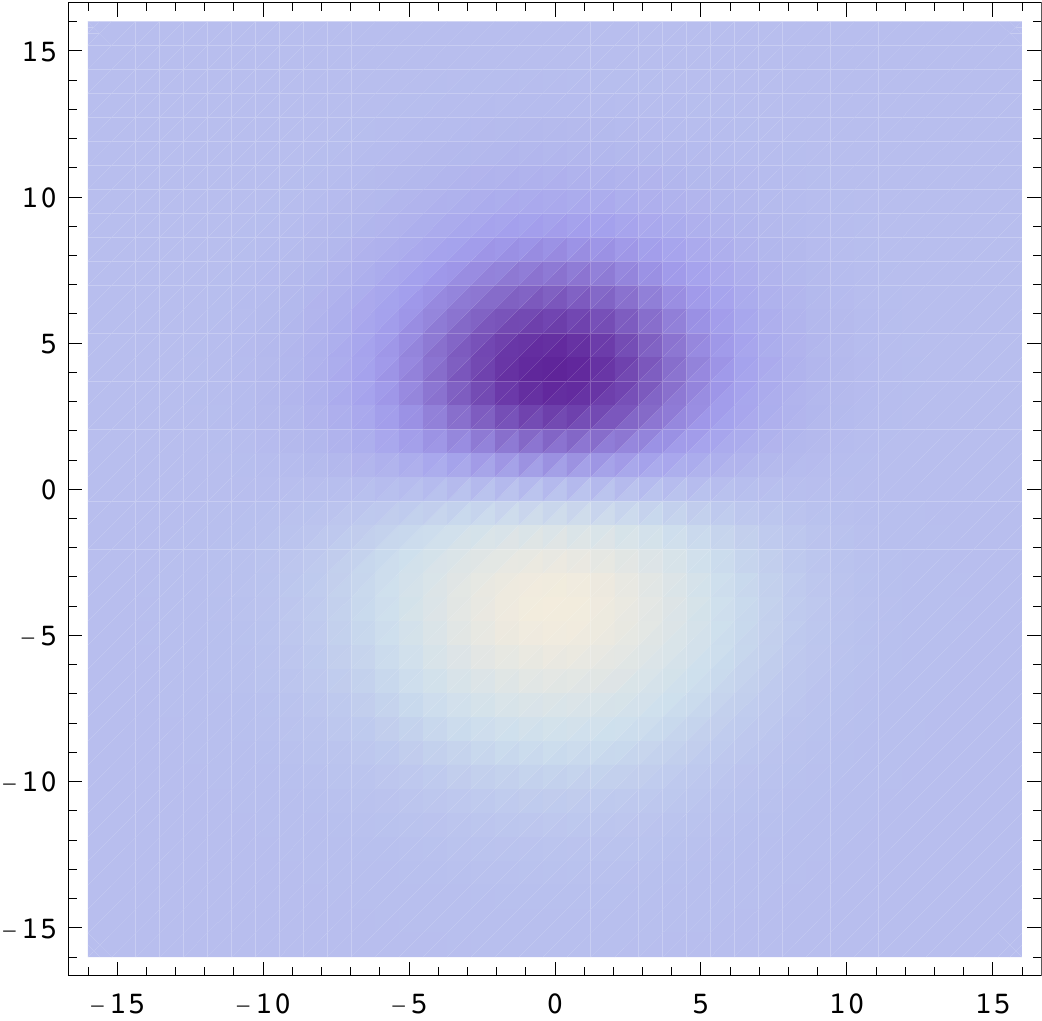} \\
    \end{tabular} 
  \end{center}
  \begin{center}
    \begin{tabular}{ccc}
      \includegraphics[width=0.14\textwidth]{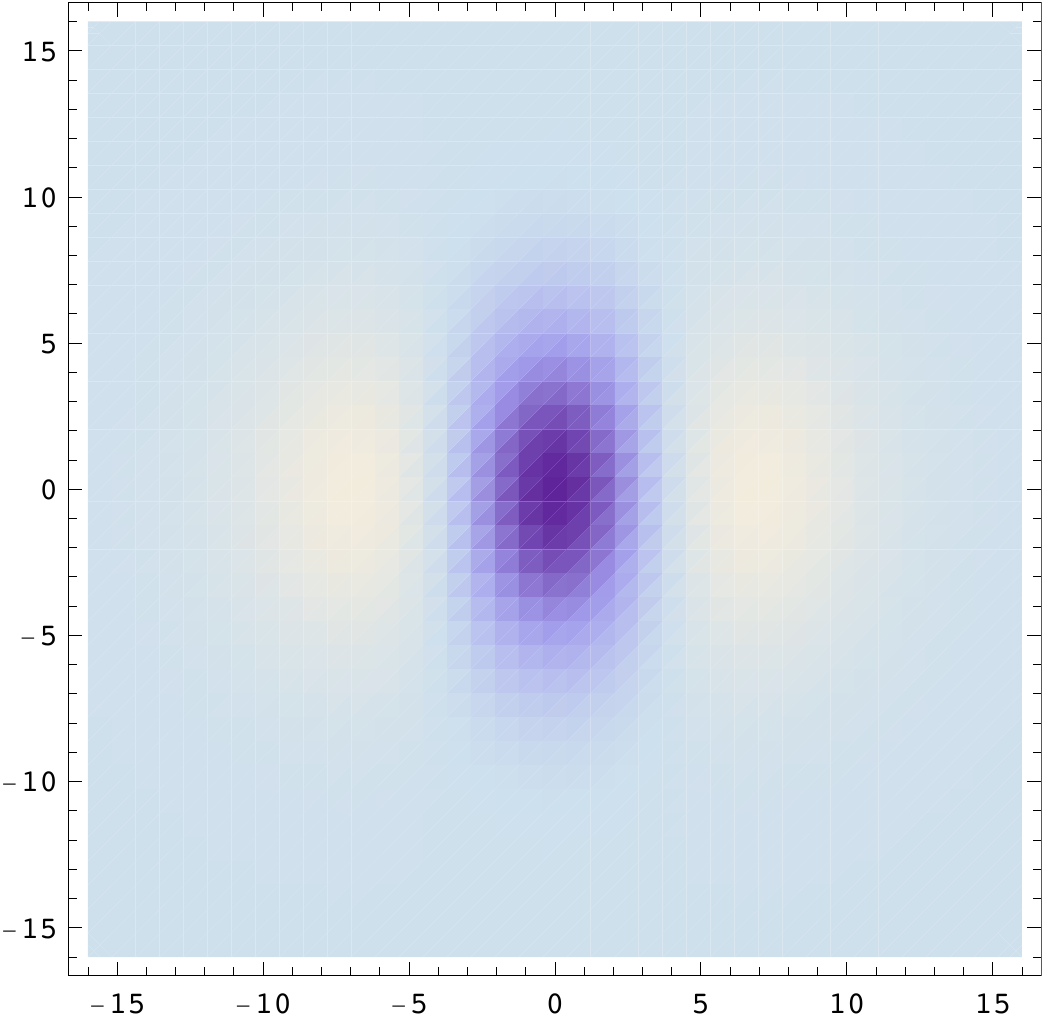} &
      \includegraphics[width=0.14\textwidth]{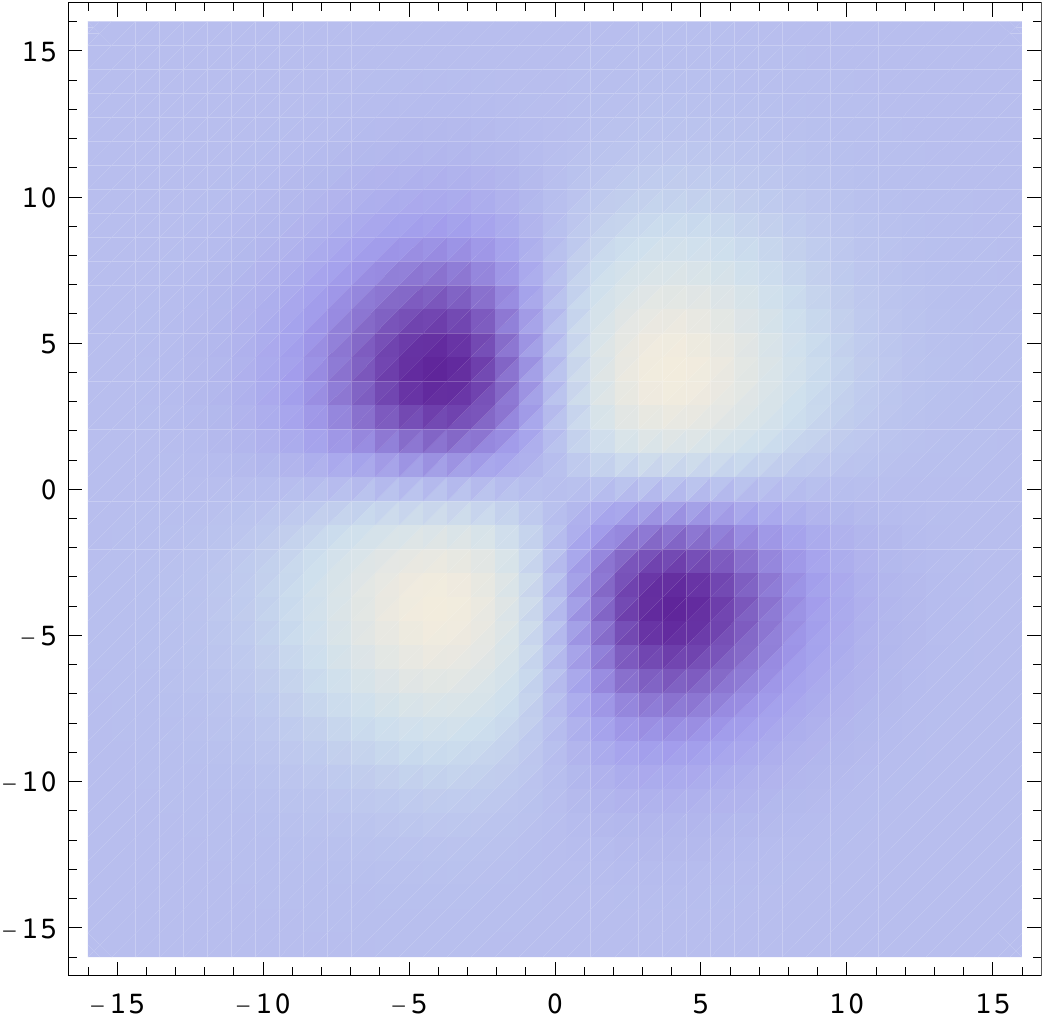} &
      \includegraphics[width=0.14\textwidth]{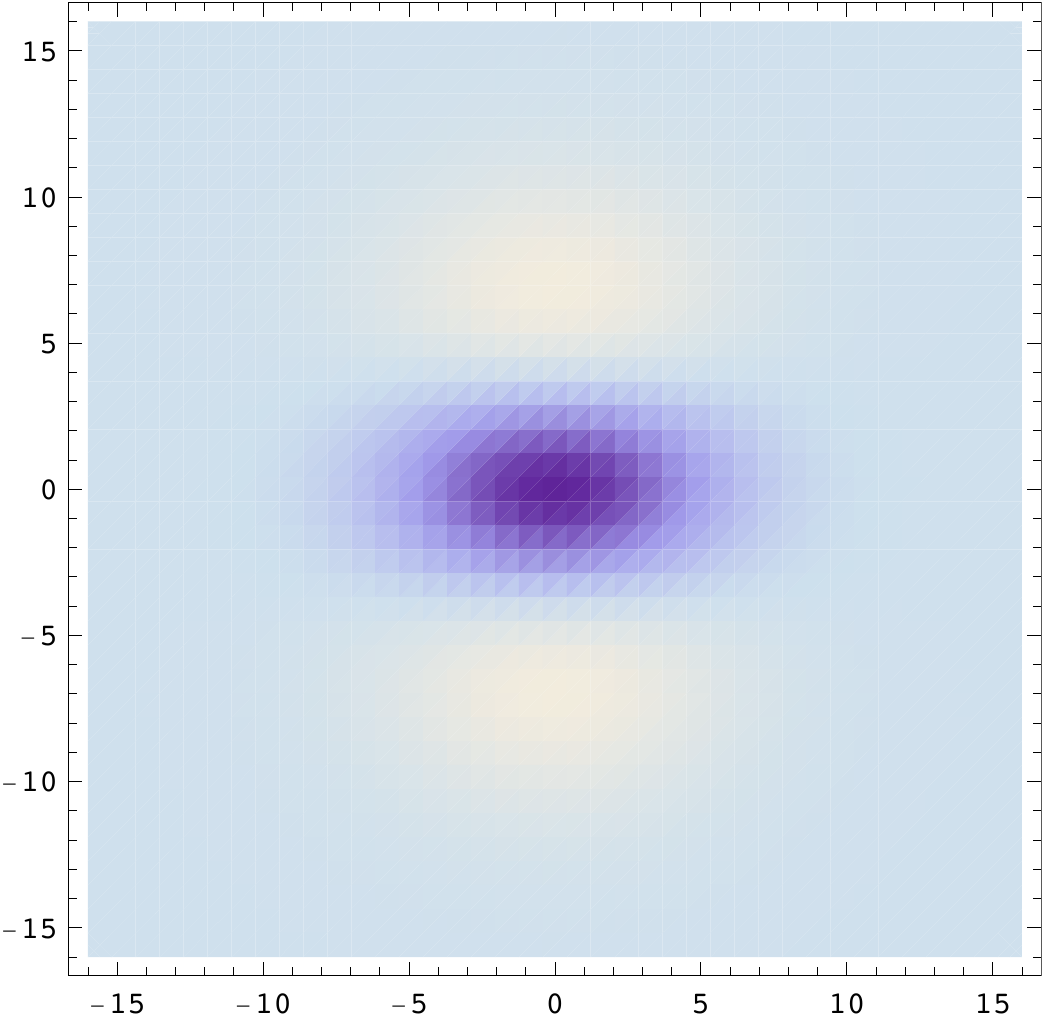} \\
    \end{tabular} 
  \end{center}
  \caption{The 2-D Gaussian kernel with its Cartesian
    partial derivatives up to order two for $\sigma = 4$.}
  \label{fig-gauss-ders}

  \begin{center}
    \begin{tabular}{cc}
      \includegraphics[width=0.14\textwidth]{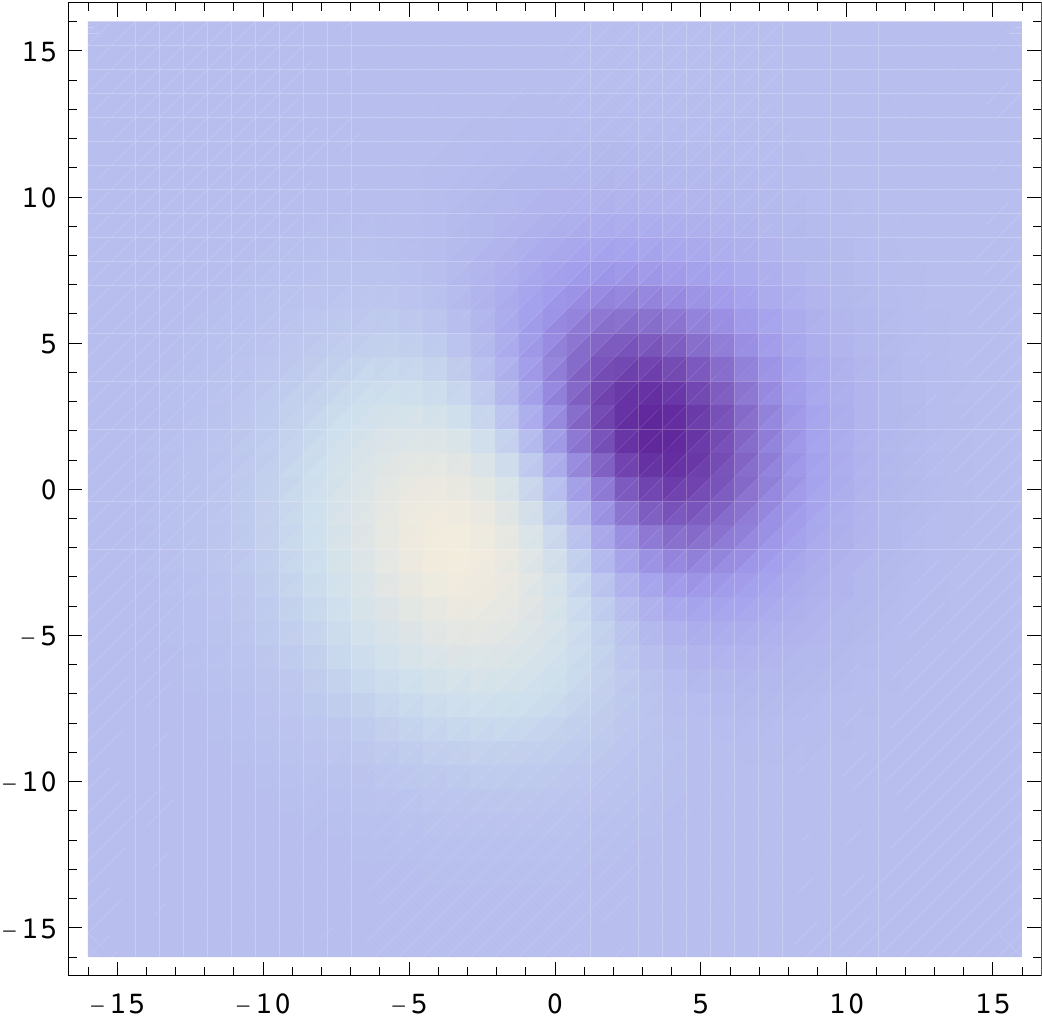} &
      \includegraphics[width=0.14\textwidth]{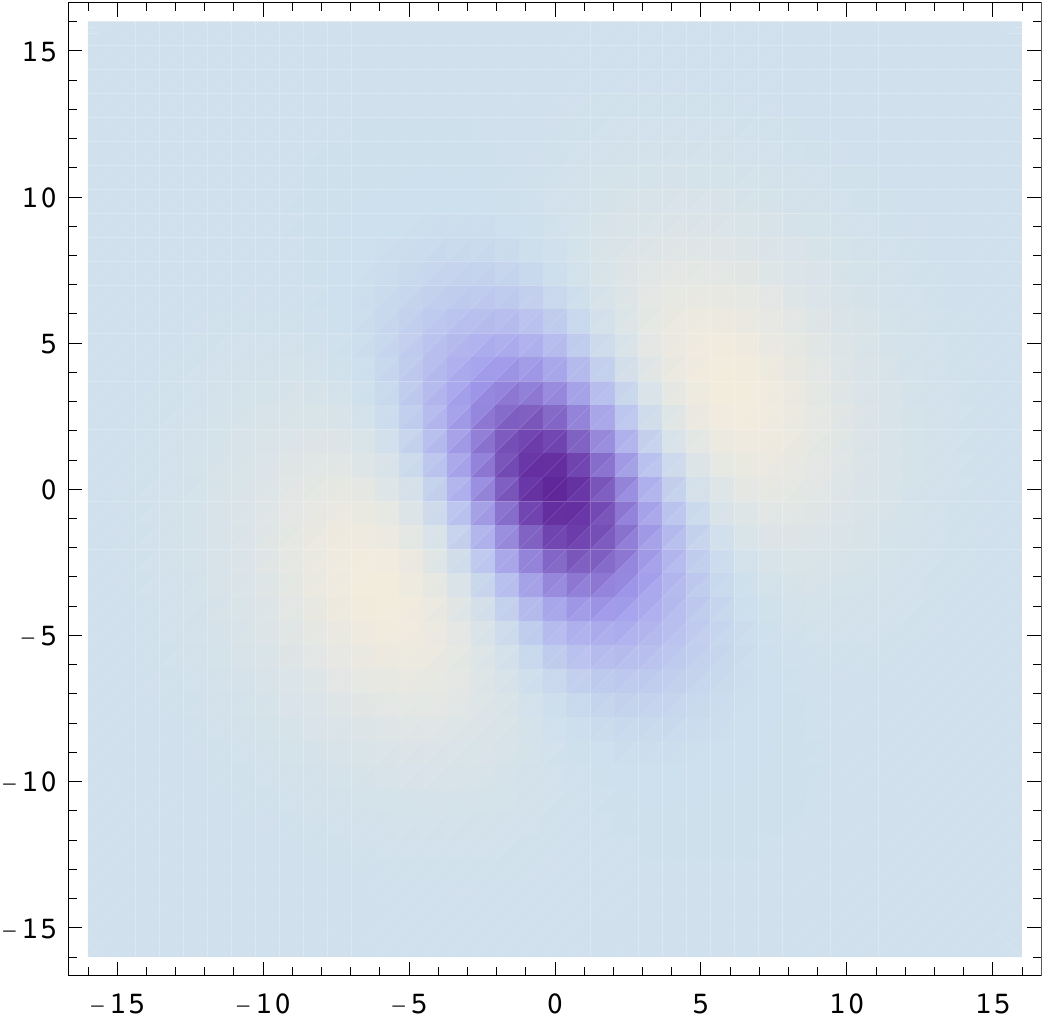} 
    \end{tabular} 
  \end{center}
  \caption{First- and second-order directional derivatives of the 2-D Gaussian kernel computed from linear combinations of
Cartesian partial derivatives according to
equations (\ref{eq-1st-order-dir-der}) and (\ref{eq-2nd-order-dir-der}) 
for $\sigma = 4$ and $\varphi=\pi/6$.}
  \label{fig-gauss-dir-ders}
\end{figure}

\subsection{Gaussian derivative layers}

Motivated by the fact that a large number of visual tasks have been
successfully addressed by first- and second-order Gaussian
derivatives, which are the primitive filters in the Gaussian 2-jet%
\footnote{The $N$-jet of a signal $f$ is the set of partial
  derivatives of $f$ up to order $N$. The Gaussian $N$-jet of a signal $f$,
  is the set of Gaussian derivatives up to order $N$, computed by
  preceding the differentiation operator by Gaussian smoothing at
  some scale (or scales).},
let us explore the consequences of using linear combinations
of first- and second-order Gaussian derivatives as the class of
possible filter weight primitives in a deep network.%
\footnote{A complementary motivation for using first- and second-order
Gaussian derivative operators as the basis for linear receptive fields, is that this is the lowest order of combining
odd and even Gaussian derivative filters. First- and second-order
derivatives naturally belong together, and serve as providing mutual
complementary information, that approximate quadrature filter pairs
\cite{KoeDoo87-BC,Lin97-IJCV,Lin18-SIIMS}.
In biological vision, approximations of first- and second-order
derivatives also occur together in the primary visual cortex
\cite{ValCotMahElfWil00-VR}.}
Thus, given an image $f$,
which could either be the input image to the deep net, or some
higher layer $F_k$ in the deep network, we first compute its scale-space
representation by smoothing with the Gaussian kernel
\begin{equation}
  \label{eq-def-sc-sp}
   L(x, y;\; \sigma) 
   = (g(\cdot, \cdot;\; \sigma) * f(\cdot, \cdot))(x, y),
\end{equation}
where 
\begin{equation}
  \label{eq-gauss-kernel}
   g(x, y;\; \sigma) 
   = \frac{1}{2 \pi \sigma^2} e^{-(x^2 + y^2)/2 \sigma^2}.
\end{equation}
Then, for simplicity with the notation for the spatial coordinates
$(x, y)$ and the scale
parameter $\sigma \in \bbbr_+$ suppressed, we consider arbitrary linear
combinations of first- and second-order Gaussian derivatives as the
class of possible linear filtering operations on $f$ (or
correspondingly for $F_k$):
\begin{multline}
  \label{eq-lin-comb-2-jet}
  J_{2,\sigma}(f) = C_0 + C_x \, L_{\xi} + C_y \, L_{\eta} \\
              + \frac{1}{2} (C_{xx} \, L_{\xi\xi} + 2 C_{xy} \, L_{\xi\eta} + C_{yy} \, L_{\eta\eta}),
\end{multline}
where $L_{\xi}$, $L_{\eta}$, $L_{\xi\xi}$, $L_{\xi\eta}$ and
$L_{\eta\eta}$ are first- and second-order 
scale-normalized derivatives \cite{Lin97-IJCV} according to 
     $L_{\xi} = \sigma \, L_x$,
     $L_{\eta} = \sigma \, L_y$,
     $L_{\xi\xi} = \sigma^{2} \, L_{xx}$,
     $L_{\xi\eta} = \sigma^{2} \, L_{xy}$ and
     $L_{\eta\eta} = \sigma^{2} \, L_{yy}$
(see Figure~\ref{fig-gauss-ders} for an illustration of Gaussian
derivative kernels)
and we have also added an offset term $C_0$ for later use in
connection with non-linearities between adjacent layers.

Based on the notion that the set of Gaussian
derivatives up to order $N$ is referred to as the Gaussian $N$-jet,
we will refer to the expression
(\ref{eq-lin-comb-2-jet}), which is to be used as computational
primitives for Gaussian derivative layers, as the {\em linearily
  combined Gaussian 2-jet\/}.

Since directional derivatives can be computed as linear
combinations of partial derivatives, for first- and second-order
derivatives we have (see Figure~\ref{fig-gauss-dir-ders} for an illustration)
\begin{align}
  \begin{split}
    \label{eq-1st-order-dir-der}
    L_{\varphi} 
     = \cos \varphi \, L_{x} + \sin \varphi \, L_{y},
  \end{split}\\ 
  \begin{split}
    \label{eq-2nd-order-dir-der}
     L_{\varphi\varphi} 
     = \cos^2 \varphi \, L_{xx} + 2 \cos \varphi \, \sin \varphi \, L_{xy} + \sin^2 \varphi \, L_{yy},
  \end{split}
\end{align}
it follows that the parameterized second-order kernels of the form
(\ref{eq-lin-comb-2-jet}) 
span all possible linear combinations of
first- and second-order directional derivatives.

The corresponding affine extension of such receptive fields, by
replacing the rotationally symmetric Gaussian kernel
(\ref{eq-gauss-kernel}) for scale-space
smoothing by a corresponding affine Gaussian kernel, does also constitute a
good idealized model for the receptive fields of simple cells in the primary visual cortex
\cite{Lin13-BICY,Lin21-Heliyon}.
In this treatment, we will, however, for simplicity restrict ourselves to regular
Gaussian derivatives, based on partial derivatives and directional
derivatives of rotationally symmetric Gaussian kernels.

In contrast to previous work in computer vision or functional
modelling of biological visual receptive fields, where Gaussian derivatives
are used as a first layer of linear receptive fields, we will, however, here
investigate the consequences of coupling such receptive fields in
cascade to form deep hierarchical image representations.

\subsection{Definition of a Gaussian derivative network}
\label{sec-def-gauss-der-nets}

To model Gaussian derivative networks with multiple feature channels
in each layer, let us assume that the input image $f$ consists of
$N_0$ image channels, for example with $N_0 = 1$ for a grey-level
image or $N_0 = 3$ for a colour image.
Assuming that each layer in the network $k \in [1, K]$ should consist of
$N_{k} \in \bbbz_+$ parallel feature channels, the feature channel $F_{1}^{c_{out}}$
with feature channel index $c_{out}  \in [1, N_1]$ in the first layer
is given by%
\footnote{Concerning the notation in this expression, please note that
  the notation for the linearily combined Gaussian $2$-jet seen in isolation
  $J_{2,\sigma_{1}}^{1,c_{out},c_{in}}(f^{c_{in}})$ should be
  understood as referring the linearily combined Gaussian $2$-jet at a single scale. When we
  in addition lift the scale parameter as a parameter argument of the
  function to the form $J_{2,\sigma_{1}}^{1,c_{out},c_{in}}(f^{c_{in}}(\cdot,
  \cdot))(x, y;\; \sigma_1)$, the resulting expression should instead then be
  understood as $\sigma_1$ being allowed to vary, as a parameter
  argument of a function.}
  \begin{equation}
  \label{def-gaussdernet-layer-1}
  F_{1}^{c_{out}}(x, y;\; \sigma_1) = 
 \sum_{c_{in} \in [1, N_0]}
  J_{2,\sigma_{1}}^{1,c_{out},c_{in}}(f^{c_{in}}(\cdot, \cdot))(x, y;\; \sigma_1), 
\end{equation}
where $J_{2,\sigma_1}^{1,c_{out},c_{in}}$ is the linearily combined
Gaussian 2-jet according to  (\ref{eq-lin-comb-2-jet}) at scale%
\footnote{For the initial theoretical treatment, we can first consider this
  representation as being defined for all scales $\sigma_1 \in \bbbr_+$. For a practical
  implementation, however, the values of $\sigma_1$ as well as for the higher
  scale levels $\sigma_k$ will be restricted to a discrete grid,
  as defined from Equations~(\ref{eq-r-geom-distr-sigma})
  and (\ref{eq-init-sc-val-disc-distr}).}
$\sigma_1$ that represents the contribution from
the image channel with index $c_{in}$ in the input image $f$ to the
feature channel with feature channel index $c_{out}$ in the first
layer and $\sigma_1$ is the scale parameter in the first layer.
The transformation between adjacent layers $k$ and $k+1$ is then for 
$k \geq 1$ given by 
\begin{multline}
  \label{def-gaussdernet-layer-k+1}
  F_{k+1}^{c_{out}}(x, y;\; \sigma_{k+1}) = \\\sum_{c_{in} \in [1, N_k]}
  J_{2,\sigma_{k+1}}^{k+1,c_{out},c_{in}}(\theta_k^{c_{in}}(F_k^{c_{in}}(\cdot, \cdot;\; \sigma_k)))(x, y;\; \sigma_{k+1}),
\end{multline}
where $J_{2,\sigma_{k+1}}^{k+1,c_{out},c_{in}}$
is the linearily combined Gaussian 2-jet according to (\ref{eq-lin-comb-2-jet}) 
that represents the contribution to the output feature
channel $F_{k+1}^{c_{out}}$  with feature channel index
$c_{out} \in [1, N_{k+1}]$ in layer $k+1$ from the input feature
channel $F_k^{c_{in}}$ with feature channel index $c_{in} \in [1, N_k]$ in
layer $k$ and $\theta_k^{c_{in}}$ represents some non-linearity, such as a ReLU
stage. The parameter $\sigma_{k+1}$ is the scale parameter for
computing layer $k + 1$ and the parameter $\sigma_k$ is the scale parameter for
computing layer $k$.

Of course, the parameters $C_0$, $C_x$, $C_y$, $C_{xx}$, $C_{xy}$ and
$C_{yy}$ in the linearily combined Gaussian 2-jet according to (\ref{eq-lin-comb-2-jet}) 
should be allowed to be
learnt differently for each layer $k \in [1, K]$ and for each combination
of input channel $c_{in}$ and output channel $c_{out}$.
Written out with explicit index notation for the layers and the input and
output feature channels for these parameters,
the explicit expression for $J_{2,\sigma_k}^{k,c_{out},c_{in}}$ in
(\ref{def-gaussdernet-layer-1}) and (\ref{def-gaussdernet-layer-k+1})
is given by
\begin{align}
  \begin{split}
  J_{2,\sigma_k}^{k,c_{out},c_{in}}(h^{c_{in}}) =
   \end{split}\nonumber\\
   \begin{split}
  = C_0^{k,c_{out},c_{in}} +
  C_x^{k,c_{out},c_{in}} \, \sigma_k \, L_{x} +
     C_y^{k,c_{out},c_{in}} \, \sigma_k \, L_{y} +
   \end{split}\nonumber\\
   \begin{split}
   + \frac{1}{2} (C_{xx}^{k,c_{out},c_{in}} \, \sigma_k^2 \, L_{xx} + 
                        2 C_{xy}^{k,c_{out},c_{in}} \, \sigma_k^2 \, L_{xy} + 
   \end{split}\nonumber\\
   \begin{split}
       \label{eq-lin-comb-2-jet-with indices}
       \hphantom{+ \frac{1}{2} (}
       + C_{yy}^{k,c_{out},c_{in}} \, \sigma_k^2 \, L_{yy}),
  \end{split}
\end{align}
with $L_x$, $L_y$, $L_{xx}$, $L_{xy}$ and $L_{yy}$ here denoting the
partial derivatives of the scale-space representation obtained by
convolving the argument $h^{c_{in}}$ with a Gaussian kernel
(\ref{eq-gauss-kernel}) with standard deviation $\sigma_k$ according
to (\ref{eq-def-sc-sp})
\begin{equation}
  L_{x^{\alpha}y^{\beta}} 
  = \partial_{x^{\alpha}y^{\beta}} 
  \left( g(\cdot, \cdot;\; \sigma_k) * h^{c_{in}}(\cdot, \cdot) \right),
\end{equation}
and with
$h^{c_{in}}$ representing the image or feature channel with index $c_{in}$
in either the input image $f$ or some higher feature layer $F_{k-1}$. 



When coupling several combined smoothing and differentiation stages in cascade in this way, with
pointwise non-linearities in between, it is natural
to let the scale parameter for layer $k$ be proportional to an initial
scale level $\sigma_0$, such that the scale parameter $\sigma_k$ in
layer $k$ is $\sigma_k = \beta_k \, \sigma_0$ for
some set of $\beta_{k} \geq \beta_{k-1} \geq 1$ and some 
minimum scale level $\sigma_0 > 0$. Specifically, it is
natural to choose the relative factors for the scale parameters according to a
geometric distribution
\begin{equation}
  \label{eq-r-geom-distr-sigma}
  \sigma_k = r^{k-1} \, \sigma_0
\end{equation}
for some $r \geq 1$.
By gradually increasing the size of the receptive fields in this way,
the transition from lower to higher layers in the hierarchy will
correspond to gradual transitions from local to regional information.
For continuous networks models, this mechanism thus provides a way 
to gradually increase the receptive field size in deeper
layers without explicit need for a discrete subsampling
operation that would imply
larger steps in the receptive fields size of deeper layers.

By additionally varying the parameter $\sigma_0$ in the above relationship, either
continuously with $\sigma_0 \in \bbbr$ or according to some self-similar distribution
\begin{equation}
  \label{eq-init-sc-val-disc-distr}
  \sigma_0 = \gamma^i
\end{equation}
for some set of integers $i \in \bbbz$ and some $\gamma > 1$,
the Gaussian derivative network defined from
(\ref{def-gaussdernet-layer-1}) and (\ref{def-gaussdernet-layer-k+1})
will constitute a multi-scale representation that will also be provably scale
covariant, as will be formally shown below in Section~\ref{sec-sc-cov}.
Specifically, the network obtained for each value of $\sigma_0$, with the derived values of
$\sigma_1$ in (\ref{def-gaussdernet-layer-1}) as well as of
$\sigma_k$ and $\sigma_{k+1}$ in (\ref{def-gaussdernet-layer-k+1}),
will be referred to as a {\em scale channel\/}.
Letting the initial scale levels be given by a self-similar
distribution in this way reflects the desire that in the absense of
further information the deep network should be agnostic with regard to
preferred scales in the input.

A similar idea of using Gaussian derivatives as structured receptive
fields in convolutional networks has also been explored in
\cite{JacGemLouSme16-CVPR}, although not in the relation to scale
covariance, multiple scale channels or using a self-similar sampling of the scale levels at
consecutive depths.

\begin{figure}[ht]
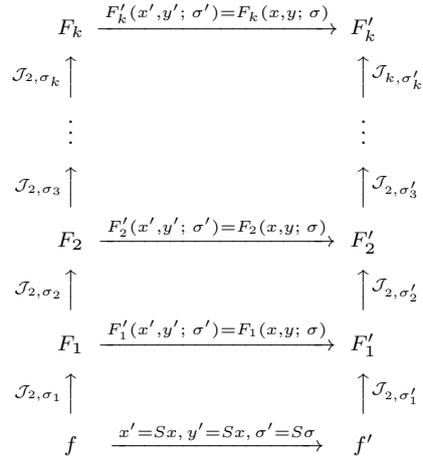

\begin{center}
    \[
      \hspace{6mm}\begin{CD}
       F_k @>{F'_k(x', y';\; \sigma') = F_k(x, y;\; \sigma)}>> F_k' \\
       @A{{\cal J}_{2,\sigma_k}}AA @AA{{\cal J}_{k,\sigma'_k}}A \\
      \vdots @. \vdots \\
        @A{{\cal J}_{2,\sigma_3}}AA @AA{{\cal J}_{2,\sigma'_3}}A \\ 
       F_2 @>{F'_2(x', y';\; \sigma') = F_2 (x, y;\; \sigma)}>> F_2' \\
        @A{{\cal J}_{2,\sigma_2}}AA @AA{{\cal J}_{2,\sigma'_2}}A \\ 
       F_1 @>{F'_1(x', y';\; \sigma') = F_1 (x, y;\; \sigma)}>> F_1' \\
        @A{{\cal J}_{2,\sigma_1}}AA @AA{{\cal J}_{2,\sigma'_1}}A \\
        f @>{x'=S x, \, y'=S x, \, \sigma' = S \sigma}>> f'
    \end{CD}
    \]
  \end{center}
  \caption{Commutative diagram for a scale-covariant Gaussian
    derivative network constructed by coupling linear combinations of
    scale-normalized Gaussian derivatives in cascade, with non-linear ReLU stages in
    between. Because of the transformation properties of the
    individual layers under scaling transformations, it will be possible to perfectly match the corresponding
    layers $F_k$ and $F_k'$ under a scaling transformation of the
    underlying image domain $f'(x') =
    f(x)$ for $x' = Sx$ and $y' = S y$, provided that
the scale parameter $\sigma_k$ in layer $k$ is proportional to the scale
parameter $\sigma_1$ in the first layer, $\sigma_k = r_k^2 \, \sigma_1$, for some
scalar constant $r_k > 1$. For such a network, the scale parameters
in the two domains should be related according to $\sigma_k' = S
\sigma_k$.
Note, however, that for a scale-discretized implementation, this commutative property
holds exactly over a continuous image domain only if the scale levels $\sigma$ and $\sigma'$ are part of the
scale grid, thus specifically only for discrete scaling factors $S$
that can be exactly represented on the discrete scale grid. For other
scaling factors, the results will instead be numerical approximations, with the
accuracy of the approximation determined by a the combination of the
network architecture with the learning algorithm.
(In this schematic illustration, we have for simplicity suppressed the
notation for multiple feature channels in the different layers, and
also suppressed the notation for the pointwise non-linearities
between adjacent layers.)}
  \label{fig-comm-diag-hier-network}
\end{figure}

\subsection{Provable scale covariance}
\label{sec-sc-cov}

To prove that a deep network constructed by coupling linear filtering
operations of the form (\ref{eq-lin-comb-2-jet}) with pointwise
non-linearities in between is scale covariant, let us consider two
images $f$ and $f'$ that are related by a scaling transformation
$f'(x', y') = f(x, y)$ for $x' = S \, x$, $y' = S \, y$ and
some spatial scaling factor $S > 0$.

A basic scale covariance property of the Gaussian scale-space representation is that
if the scale parameters $\sigma$ and $\sigma'$ in the two image domains are
related according to $\sigma' = S \, \sigma$, then the Gaussian
scale-space representations are equal at matching image points and
scales \cite[Eq.~(16)]{Lin97-IJCV} 
\begin{equation}
   L'(x', y';\; \sigma') = L(x, y;\; \sigma).
\end{equation}
Additionally regarding spatial derivatives, it follows from a
general result in \cite[Eq.~(20)]{Lin97-IJCV} that the
scale-normalized spatial derivatives will also be equal (when using scale
normalization power $\gamma = 1$ in
the scale-normalized derivative concept):
\begin{equation}
  L'_{\xi^{\alpha}\eta^{\beta}}(x', y';\; \sigma') = L_{\xi^{\alpha}\eta^{\beta}}(x, y;\; \sigma).
\end{equation}
Applied to the linearily combined Gaussian 2-jet in (\ref{eq-lin-comb-2-jet}),
it follows that also the linearily combined Gaussian 2-jets computed from two
mutually rescaled scalar image patterns $f$ and $f'$ will be related by
a similar scaling transformation
\begin{equation}
  J_{2,\sigma'}(f'(\cdot, \cdot)) (x', y';\; \sigma') 
 = J_{2,\sigma}(f(\cdot, \cdot)) (x, y;\; \sigma) 
\end{equation}
provided that the image positions $(x, y)$ and $(x', y')$ and the 
scale parameters $\sigma$ and $\sigma'$ are appropriately matched, 
and that the parameters $C_0$, $C_x$, $C_y$, $C_{xx}$, $C_{xy}$ and
$C_{yy}$ in (\ref{eq-lin-comb-2-jet}) are shared between the two
2-jets at the different scales.

Thus, provided that the image positions and the scale levels are
appropriately matched according to $x' = S \, x$, $y' = S \, y$ and $\sigma' = S \,
\sigma$, it holds that the corresponding feature channels $F_1^{c_{out}}$ and ${F'}_1^{c_{out}}$ in
the first layer according to (\ref{def-gaussdernet-layer-1}) are equal
up to a scaling transformation:
\begin{align}
  \begin{split}
     {F'}_1^{c_{out}} (x', y';\; \sigma'_1)
  \end{split}\nonumber\\
  \begin{split}
   = \sum_{c_{in} \in [1, N_0]} J_{2,\sigma'_1}^{1,c_{out},c_{in}}({f'}^{c_{in}}(\cdot, \cdot)) (x', y';\; \sigma'_1) = \\
   \end{split}\nonumber\\
  \begin{split}
 =\sum_{c_{in} \in [1, N_0]}  J_{2,\sigma_1}^{1,c_{out},c_{in}}(f^{c_{in}}(\cdot, \cdot)) (x, y;\; \sigma_1) 
   \end{split}\nonumber\\
  \begin{split}
     \label{eq-first-layer}
 = F_1^{c_{out}}(x, y;\; \sigma_1),
  \end{split}
\end{align}
provided that the parameters $C_0^{1,c_{out},c_{in}}$,
$C_x^{1,c_{out},c_{in}}$, \\ $C_y^{1,c_{out},c_{in}}$, 
$C_{xx}^{1,c_{out},c_{in}}$, $C_{xy}^{1,c_{out},c_{in}}$ and $C_{yy}^{1,c_{out},c_{in}}$ 
in the 2-jets according to  (\ref{eq-lin-comb-2-jet-with indices}) for
$k = 1$ are
shared between the scale channels for corresponding combinations of feature channels, 
as indexed by $c_{in}$ and $c_{out}$.

By continuing a corresponding construction of higher layers, by
applying similar operations in cascade, with pointwise non-linearities
such as ReLU stages in between, 
according to (\ref{def-gaussdernet-layer-k+1})
with the initial scale levels
$\sigma_0$ and $\sigma_0'$ in (\ref{eq-r-geom-distr-sigma}) related
according to $\sigma_0' = S \, \sigma_0$, it follows that also
the corresponding feature channels ${F'}_{k+1}^{c_{out}}$ and
$F_{k+1}^{c_{out}}$ in the higher layers in the hierarchy 
will be equal up to a scaling transformation, since the input data
from the adjacent finer layers ${F'}_{k}^{c_{in}}$ and
$F_{k}^{c_{in}}$  are already known to be related by plain scaling
transformation (see Figure~\ref{fig-comm-diag-hier-network} for an illustration):
\begin{align}
  \begin{split}
   {F'}_{k+1}^{c_{out}}(x', y';\; \sigma'_{k+1}) 
  \end{split}\nonumber\\
  \begin{split}
  = \sum_{c_{in} \in [1, N_k]}
         J_{2,\sigma'_{k+1}}^{k+1,c_{out},c_{in}}(\theta_k^{c_{in}}({F'}_k^{c_{in}}(\cdot, \cdot;\; \sigma'_k)))(x', y';\; \sigma'_{k+1}) 
  \end{split}\nonumber\\
  \begin{split}
  =  \sum_{c_{in} \in [1, N_k]} J_{2,\sigma_{k+1}}^{k+1,c_{out},c_{in}}(\theta_k^{c_{in}}(F_k^{c_{in}}(\cdot, \cdot;\; \sigma_k)))(x, y;\; \sigma_{k+1})
   \end{split}\nonumber\\
  \begin{split}
    = F_{k+1}^{c_{out}}(x, y;\; \sigma_{k+1}),
  \end{split}
\end{align}
again assuming that the parameters $C_0^{k,c_{out},c_{in}}$, $C_x^{k,c_{out},c_{in}}$, $C_y^{k,c_{out},c_{in}}$, 
$C_{xx}^{k,c_{out},c_{in}}$, $C_{xy}^{k,c_{out},c_{in}}$ and $C_{yy}^{k,c_{out},c_{in}}$ 
in the 2-jets according to (\ref{eq-lin-comb-2-jet-with indices}) are shared between the scale
channels for corresponding layers and combinations of feature channels,
as indexed by $c_{in}$ and $c_{out}$.

A pointwise non-linearity, such as a ReLU stage, trivially
commutes with scaling transformations and does therefore not affect the
scale covariance properties.

In a recursive manner, we thereby prove that scale covariance in
the lower layers imply scale covariance in any higher layer.

\subsection{Provable scale invariance}
\label{sec-sc-inv}

To prove scale invariance after max pooling over the scale channels, let us assume that we
have an {\em infinite\/} set of scale channels ${\cal S}$, with the initial
scale values $\sigma_0$ in (\ref{eq-init-sc-val-disc-distr}) either continuously distributed with
\begin{equation}
  {\cal S} = \{ \sigma \in \bbbr_+ \}
\end{equation}
or discrete in the set
\begin{equation}
  {\cal S} = \{ \sigma_i = \gamma^i,
  \forall i \in \bbbz \}.
\end{equation}
for some $\gamma > 1$.

The max pooling operation over the scale channels for feature channel
$c$ in layer $k$ will then at every image position 
$(x, y)$ and with the scale parameter in this layer $\sigma_k$ and the
relative scale factor $r$ according to (\ref{eq-r-geom-distr-sigma}) return the value
\begin{equation}
  F_{k,sup}^c(x, y) 
  = \sup_{\sigma_k}  F_k^c(x, y;\; \sigma_k) 
  = \sup_{\sigma_0 \in {\cal S}} F_k^c(x, y;\; r^{k-1} \sigma_0).
\end{equation}
Let us without essential loss of generality assume that the scaling
transformation is performed around the image point $(x, y)$,
implying that we can without essential loss of generality assume that
the origin is located at this point. The set of feature values before the
scaling transformation is then given by
\begin{equation}
   M = \{ F_k^c(0, 0;\; r^{k-1} \sigma_0) \, \forall \sigma_0 \}
\end{equation}
and the set of feature values after the scaling transformation
\begin{equation}
   M' = \{ {F'}^c_k(0, 0;\; r^{k-1} \sigma_0') \, \forall \sigma_0' \}.
\end{equation}
With the relationship $\sigma_0' = S \, \sigma_0$, these sets are clearly
equal provided that $S > 0$ in the continuous case or $S = \gamma^j$
for some $j \in \bbbz$ in the discrete case, implying that the
supremum of the set is preserved (or the result of any other
permutation invariant pooling operation, such as the average).
Because of the closedness property under scaling transformations,
the scaling transformation just shifts the set of feature
values over scales along the scale axis.

In this way, the result
after max pooling over the scale channels is essentially scale invariant, in the sense that
the result of the max pooling operation at any image point follows the
geometric transformation of the image point under scaling
transformations.

When using a {\em finite\/} number of scale channels, the result of
max pooling over the scale channels is, however, not guaranteed to truly scale invariant, since there could be
boundary effects, implying that the maximum over scales moves in to or
out from a finite scale interval, because of the scaling
transformation that shifts the position of the scale maxima on the
scale axis.

To reduce the likelihood of such effects occurring, we propose as design
criterion to
ensure that there should be a sufficient number of additional scale
channels below and above the effective training scales%
\footnote{To formally define the notion of ``effective training
  scale'', we can consider the set of scale values of the scale channels
  that lead to the maximum value over scales for a max pooling network
  (or the range of scales that contain the dominant mass over scales
  for an average pooling network), formed as the union of all the
  samples in the training data of a specific size, and with some
  additional cutoff function to select the majority of the responses,
  specifically with some
  suppression of spurious outliers. Since these resulting scale levels
  could be expected to vary between different training samples of
  roughly the same size, the notion of ``effective training scales''
  should be a scale interval rather than a single scale, and may vary
  depending upon the properties of the training data.}
in the training data.
The intention behind this is that the
learning algorithm could then learn that the image structures that occur
below and above 
the effective training scales are less relevant, and thereby associate lower values of
the feature maps to such image structures. In this way, the risk should be reduced
that erroneous types of image structures are being picked up by
the max pooling operation over the multiple scale channels, by the
scale channels near the scale boundaries thereby having lower magnitude values
in their feature maps than the central ones.

A similar scale boundary handling strategy is used in the scale channel networks based
on applying a fixed CNN to a set of rescalings of the original image
in \cite{JanLin21-ICPR}, as opposed to the approach here based on applying a set of rescaled CNNs
to a fixed size input image.

\section{Experiments with a single-scale-channel network}
\label{sec-exp-mnist}

To investigate the ability of these types of deep hierarchical Gaussian derivative networks to
capture image structures with different image shapes, we first did initial experiments with the
regular MNIST dataset \cite{LecBotBenHaf98-ProcIEEE}.
We constructed a 6-layer network in PyTorch \cite{PasGroChiChaYanDeVLinDesAntLer17-NIPS}
with 12-14-16-20-64 channels in the
intermediate layers and 10 output channels, intended to learn each type of digit.

We chose the initial scale level $\sigma_0 = 0.9$~pixels and the
relative scale ratio $r = 1.25$ in (\ref{eq-r-geom-distr-sigma}), implying that the maximum
value of $\sigma$ is $0.9 \times 1.25^5 \approx 2.7$~pixels relative to the image size of $28
\times 28$ pixels. The individual receptive fields do then have
larger spatial extent because of the spatial extent of the Gaussian
kernels used for image smoothing and the larger positive and negative side lobes
of the first- and second-order derivatives.

We used regular ReLU stages between the filtering steps, but no
spatial max
pooling or spatial stride, and no fully connected layer,
since such operations would destroy the scale covariance.
Instead, the receptive fields are solely determined from linear
combinations of Gaussian derivatives, with successively larger 
receptive fields of size $\sigma_0 \, r^{k-1}$, which enable a gradual
integration from local to regional image structures.
In the final layer, only the value at the central pixel is extracted,
or here for an even image size, the average over the central $2 \times
2$ neighbourhood, which, however, destroys full scale covariance.
To ensure full scale covariance, the input images should instead have
odd image size.

\begin{figure}[t]
  \begin{center}
    \begin{tabular}{c}
       \includegraphics[width=0.48\textwidth]{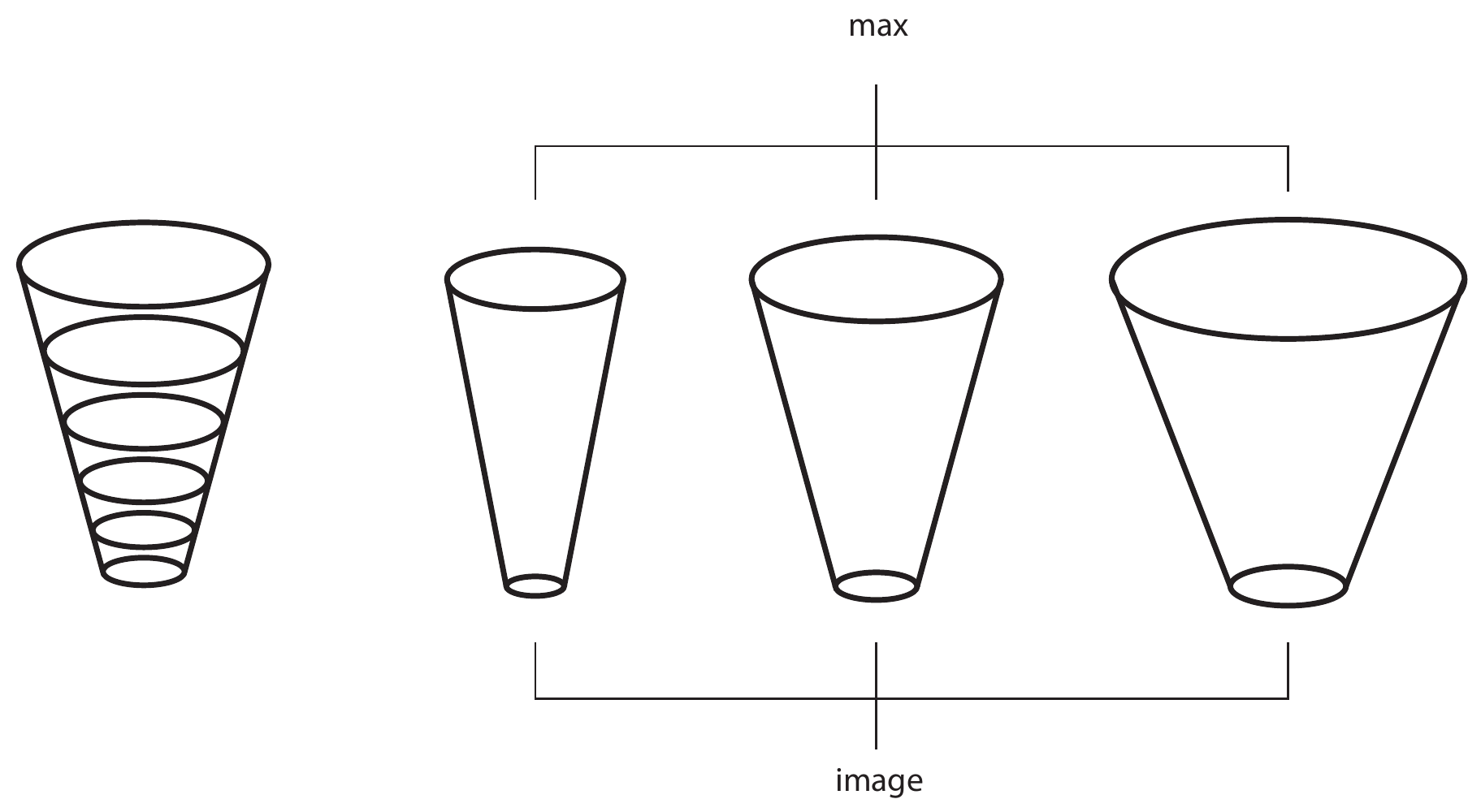}
    \end{tabular}
  \end{center}
\caption{(left) Schematic illustration of the architecture of the
  single-scale-channel network, with 6 layers of receptive fields at
  successively coarser levels of scale. (right) Schematic illustration of the
  architecture of a multi-scale-channel network, with multiple parallel scale
  channels over a self-similar distribution of the initial scale level $\sigma_0$
  in the hierarchy of Gaussian derivative layers coupled in cascade.}
\label{fig-architecture}
\end{figure}

\begin{figure*}[hbt]
  \begin{center}
    \begin{tabular}{c}
       \includegraphics[width=0.80\textwidth]{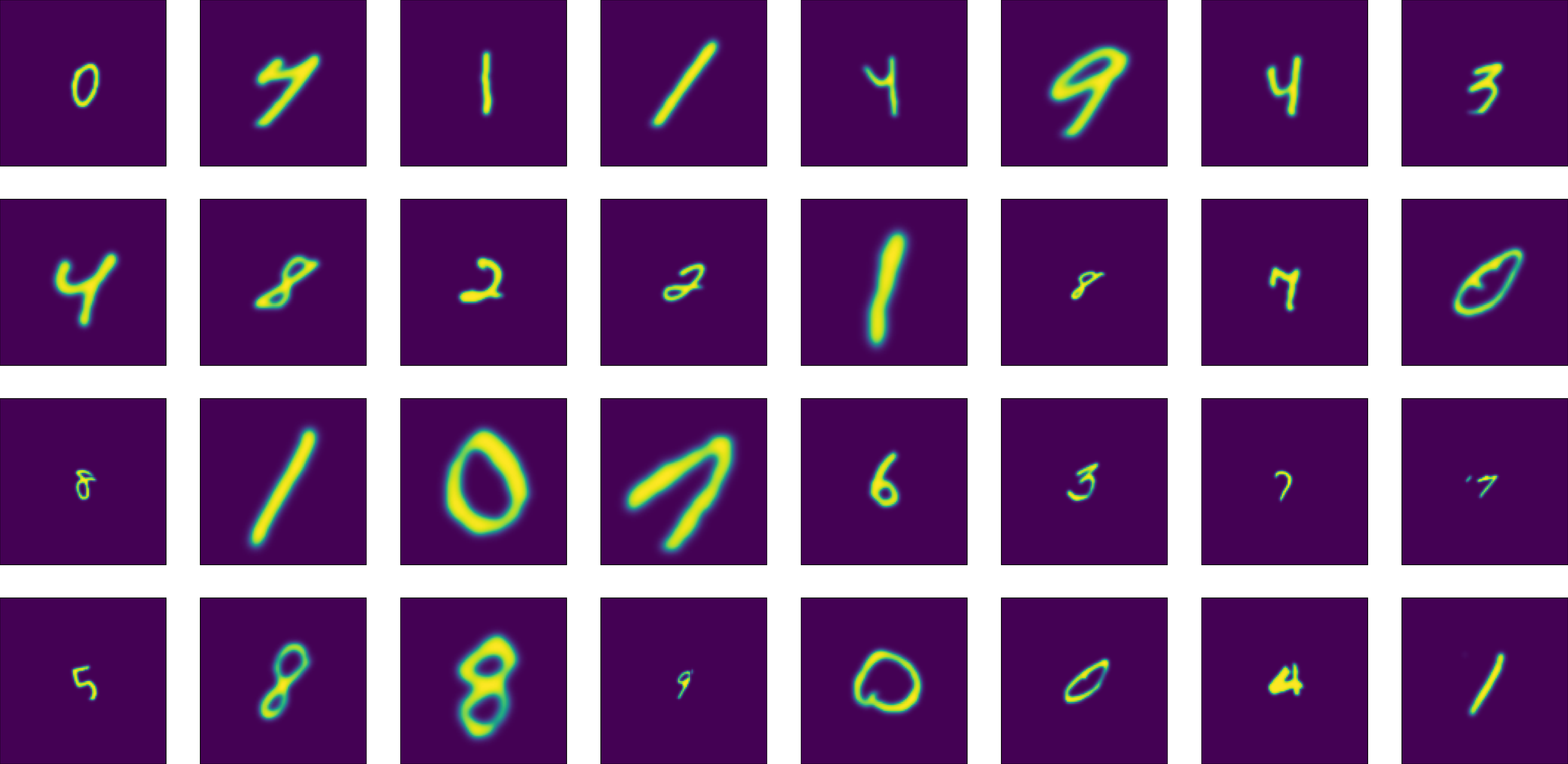}
    \end{tabular}
  \end{center}
\caption{Sample images from the MNIST Large Scale dataset \cite{JanLin21-ICPR,JanLin20-MNISTLargeScale}. This figure
  shows digits for sizes in the range [1, 4], for which there are
  training data. In addition, the MNIST Large Scale dataset contains
  testing data over the wider size range [1/2, 8].}
  \label{fig-ill-MNISTLargeScale}
\end{figure*}

The network was trained on 50 000 of the training images in the dataset,
with the offset term $C_0^{k,c_{out},c_{in}}$ that serves as the bias
for the non-linearity and the weights $C_x^{k,c_{out},c_{in}}$, $C_y^{k,c_{out},c_{in}}$, 
$C_{xx}^{k,c_{out},c_{in}}$, $C_{xy}^{k,c_{out},c_{in}}$ and $C_{yy}^{k,c_{out},c_{in}}$ 
of the Gaussian derivatives in (\ref{eq-lin-comb-2-jet-with indices})
initiated to random values and trained individually for each layer and
feature channel by stochastic gradient descent over 40 epochs
using the Adam optimizer \cite{KinBa15-ICLR} set to minimize the
binary cross-entropy loss. We used a cosine learning curve with maximum
learning rate of 0.01 and minimum learning rate 0.00005 and
using batch normalization over batches with 50 images.
The experiment lead to 99.93~\% training accuracy and 99.43~\% test
accuracy on the test dataset containing 10 000 images.

Notably, the training accuracy does not reach 100.00~\%, probably because
of the restricted shapes of the filter weights, as determined by the 
{\em a priori\/} shapes of the receptive fields in terms of linear
combinations of first- and second-order Gaussian derivatives. Nevertheless, the
test accuracy is quite good given the moderate number of parameters in
the network ($6 \times (12 + 12 \times 14 + 14 \times 16 + 16 \times
20 + 20 \times 64 + 64 \times 10) = 15~864$).

\subsection{Discrete implementation}

In the numerical implementation of scale-space smoothing, we used
separable smoothing with the discrete analogue of the Gaussian kernel
$T(n;\; s) = e^{-s} I_n(s)$ for $s = \sigma^2$ in terms of
modified Bessel functions $I_n$ of integer order \cite{Lin90-PAMI}.%
\footnote{This way of implementing Gaussian convolution on discrete spatial domain
                 corresponds to the solution of a purely spatial
                 discretization of the diffusion equation
                 $\partial_s L = \frac{1}{2} \nabla^2 L$ that
                 describes the effect of Gaussian convolution
                 (\ref{eq-def-sc-sp}), with the continuous Laplacian
                 operator $\nabla^2$ replaced by the five-point operator
                $\nabla_5^2$ defined by $(\nabla_5^2 L)(x, y) = L(x+1, y) + L(x-1, y) + L(x,
                y+1) + L(x, y-1) - 4 L(x, y)$ \cite{Lin90-PAMI}.}
The discrete derivative approximations were computed by central
differences, $\delta_x = (-1/2, 0, 1/2)$, $\delta_{xx} = (1, -2,
1)$, $\delta_{xy} = \delta_x \delta_y$, etc., implying that the spatial smoothing operation can be
shared between derivatives of different order, and implying that
scale-space properties are preserved in the discrete implementation
of the Gaussian derivatives \cite{Lin93-JMIV}.
This is a general methodology for computing Gaussian derivatives for a
large number of visual tasks.

By computing the Gaussian derivative responses in this way, the
scale-space smoothing is only performed once for each scale level, and there
is no need for repeating the scale-space smoothing for each order of
the Gaussian derivatives.

\section{Experiments with a multi-scale-channel network}
\label{sec-exp-mnist-large-scale}

To investigate the ability of a multi-scale-channel network to handle spatial scaling transformations, we
made experiments on the MNIST Large Scale dataset
\cite{JanLin21-ICPR,JanLin20-MNISTLargeScale}.
This dataset contains rescaled digits from the original MNIST
dataset \cite{LecBotBenHaf98-ProcIEEE} embedded in images of size $112
\times 112$, see Figure~\ref{fig-ill-MNISTLargeScale} for an
illustration.

For training, we used either of the datasets containing 50~000 rescaled digits with
relative scale factors 1, 2 or 4,
respectively, henceforth referred to as training sizes 1, 2 and 4.
For testing, the dataset contains 10~000 rescaled digits with relative
scale
factors between 1/2 and 8, respectively, with a relative scale ratio of $\sqrt[4]{2}$
between adjacent testing sizes.

\begin{figure*}[hbtp]
  \begin{center}
    \begin{tabular}{c}
       {\em Scale generalization performance when training on size 1} \\
       \includegraphics[width=0.70\textwidth]{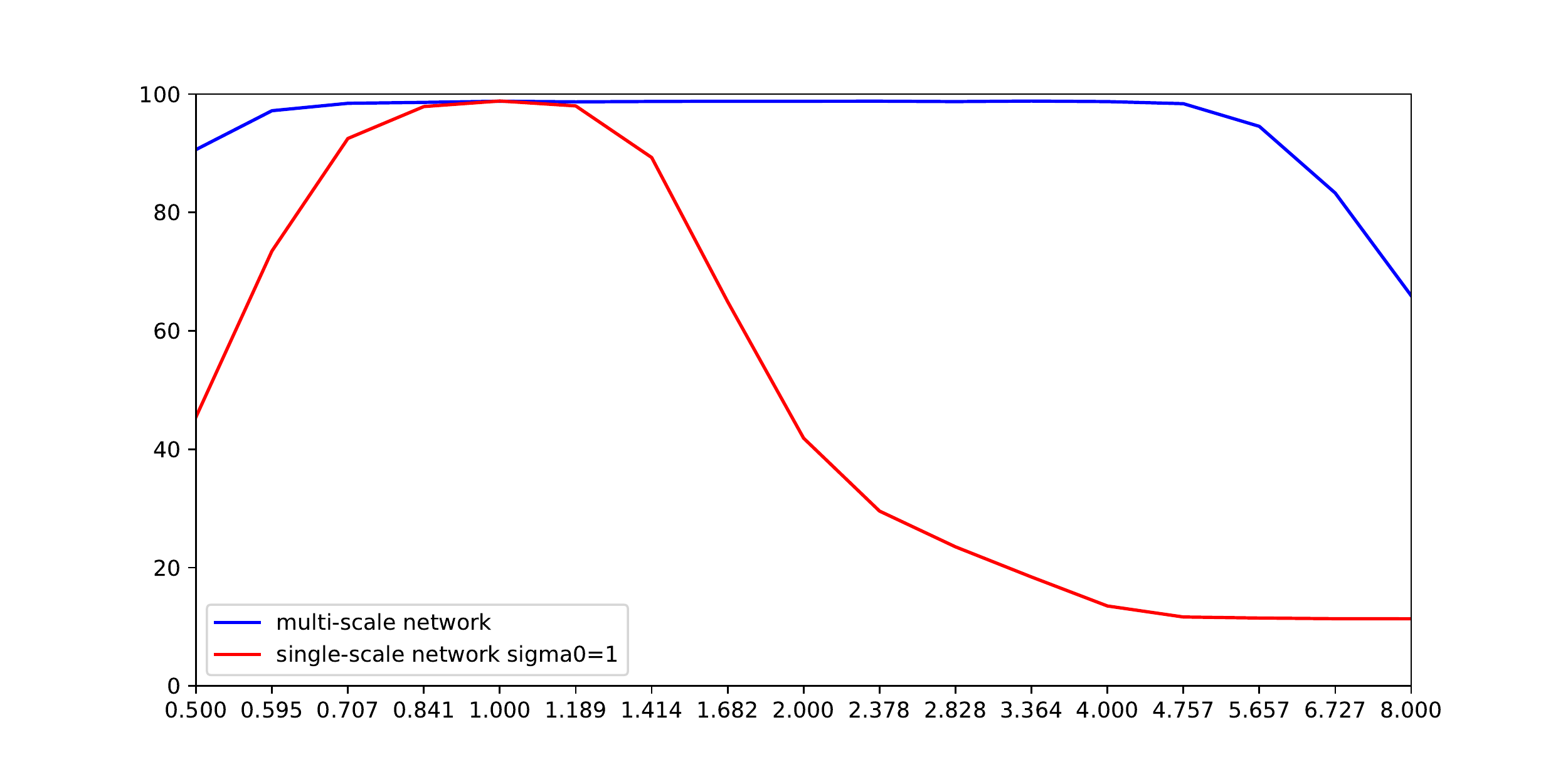}
      \\
       {\em Scale generalization performance when training on size 2} \\
       \includegraphics[width=0.70\textwidth]{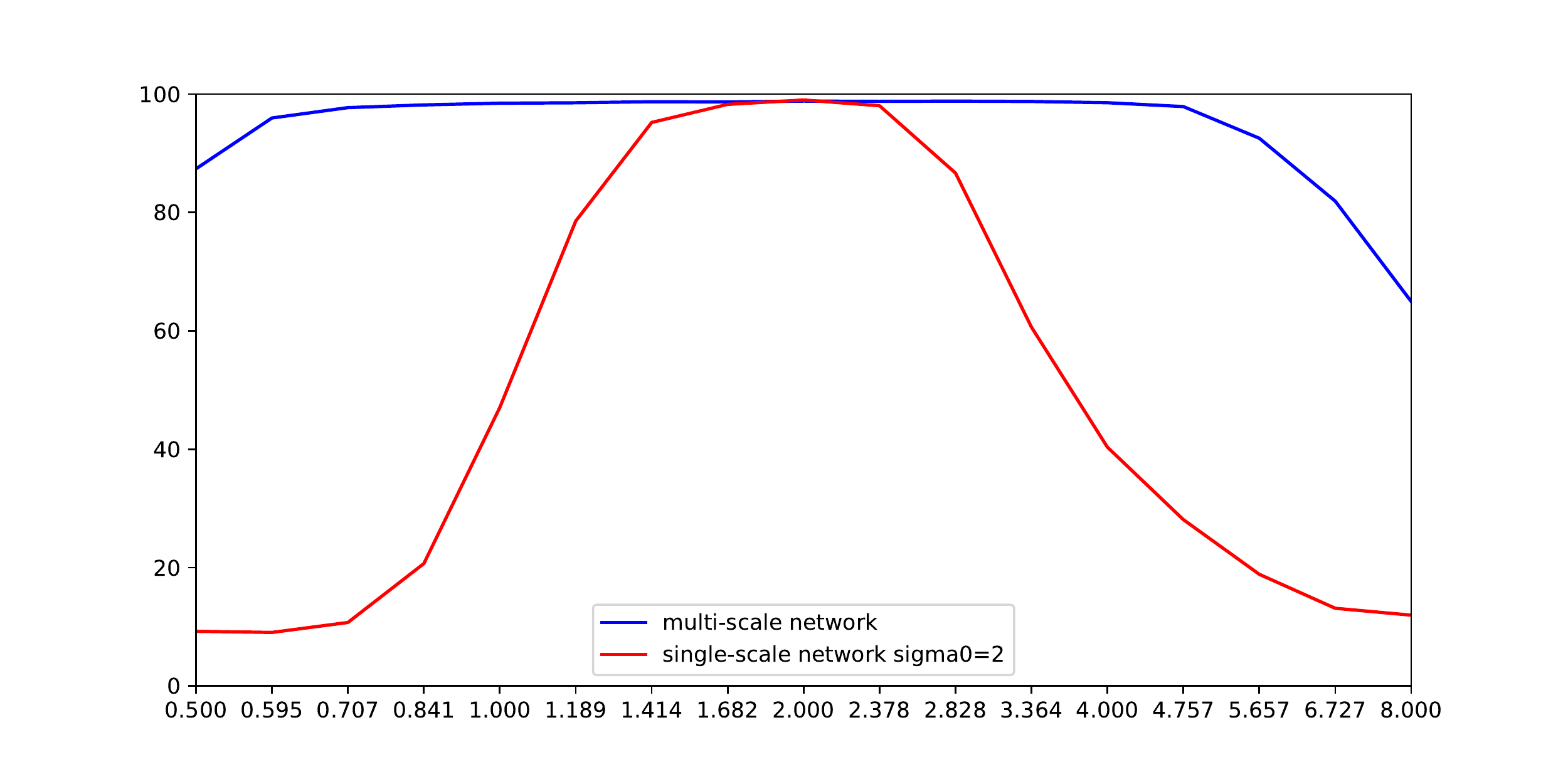}
      \\
       {\em Scale generalization performance when training on size 4} \\
       \includegraphics[width=0.70\textwidth]{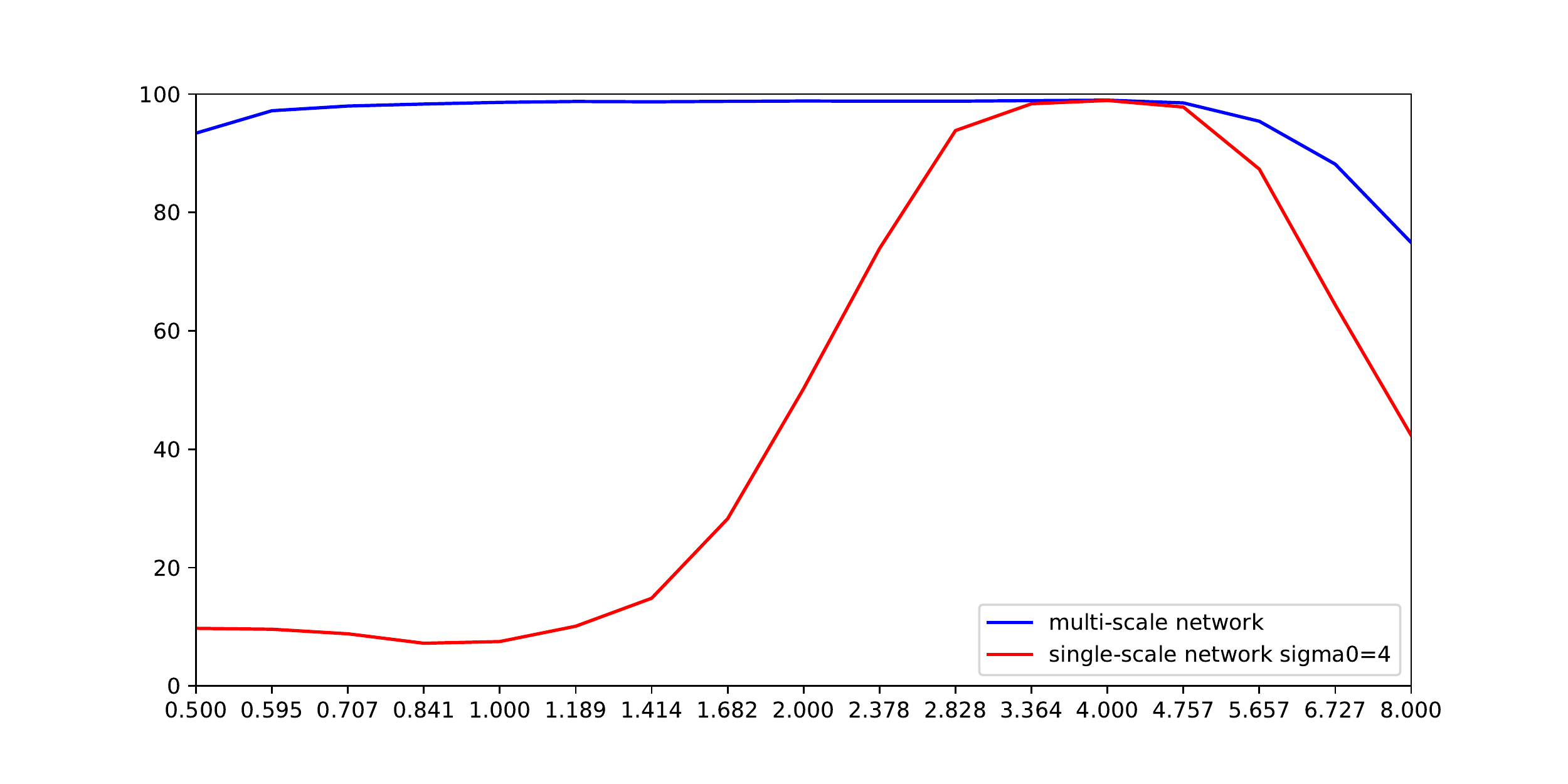}  \\
    \end{tabular}
  \end{center}
  \caption{Experiments showing the ability of a multi-scale-channel
    network to generalize to new scale levels not present in the
    training data. In the top row, all training data are for size
    1, whereas we evaluate on multiple testing sets for each one of the sizes between 1/2 and 8.
    The red curve shows the generalization performance for a
    single-scale-channel network for $\sigma_0 = 1$, whereas the blue curve shows the
    result for a multi-scale-channel network covering the range of
    $\sigma_0$-values between $1/\sqrt{2}$ and 8. As can be seen from
    the result, the generalization ability is much better for the
    multi-scale-channel network compared to the single-scale-channel network.
    In the middle row, a similar type of experiment is repeated for
    training size 2 and with $\sigma_0= 2$ for the single-scale-channel network.
    In the bottom row, a similar experiment is performed for training
    size 4 and with $\sigma_0= 4$ for the single-scale-channel network.
(Horizontal axis: Scale of testing data.)}
\label{fig-generalization-exp-sizes-1-2-4}
\end{figure*}

 \begin{figure*}[hbt]
   \begin{center}
     \begin{tabular}{c}
        {\em Scale generalization performance when training on
       different sizes: Default network} \\
        \includegraphics[width=0.80\textwidth]{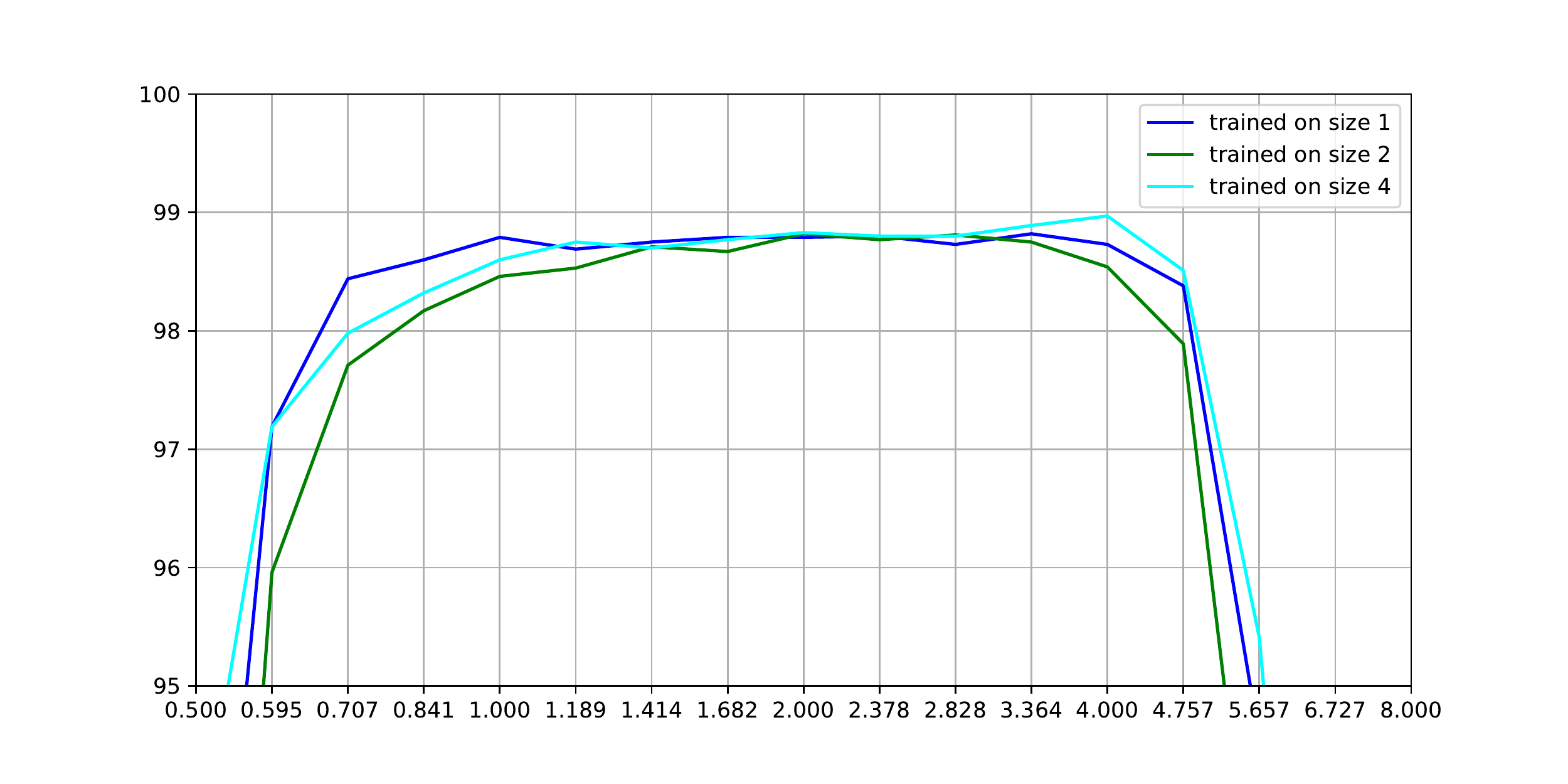}
       \\
        {\em Scale generalization performance when training on different sizes: Larger network} \\
        \includegraphics[width=0.80\textwidth]{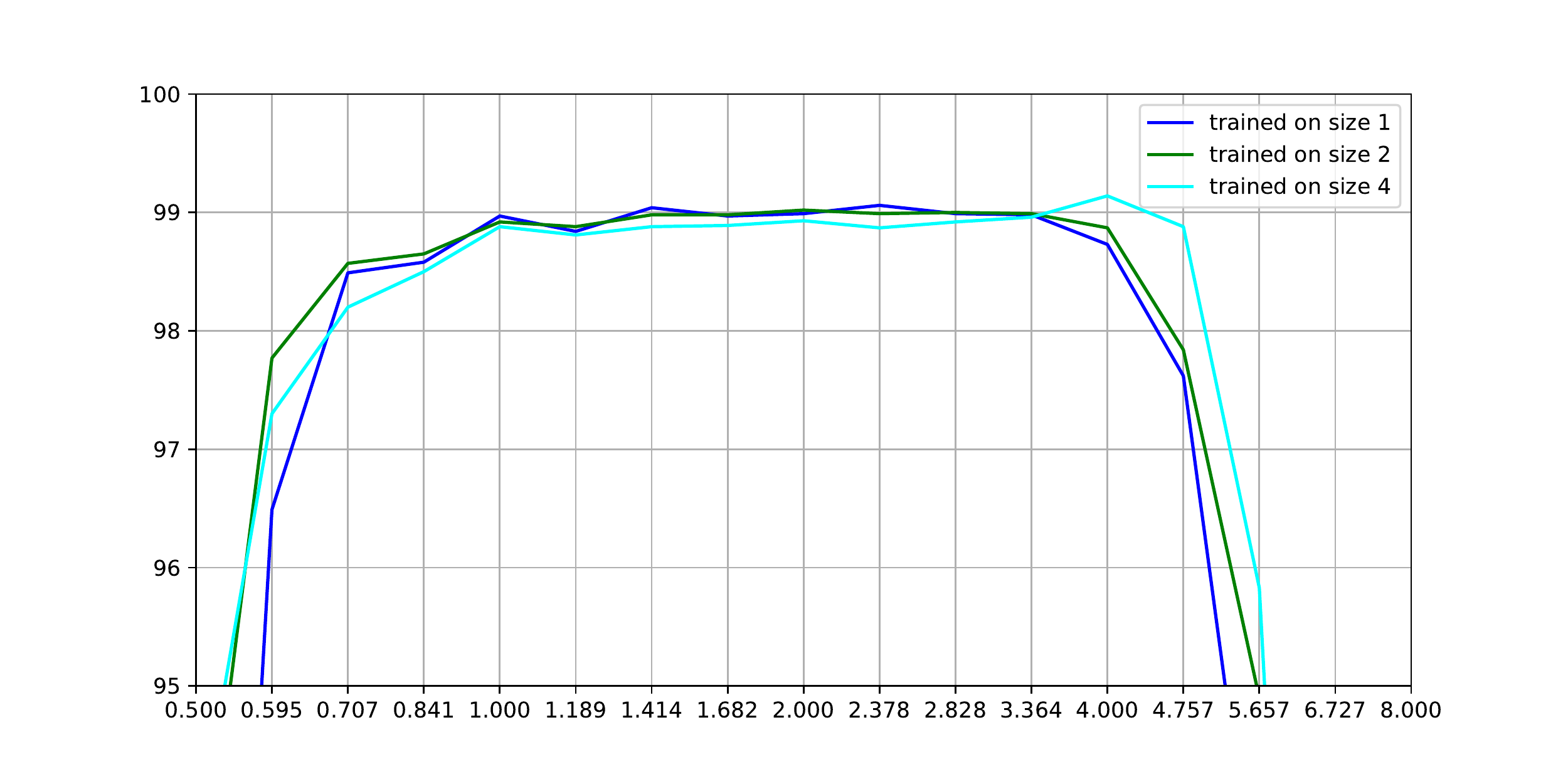}
     \end{tabular}
   \end{center}
   \caption{Joint visualization of the generalization performance when
    training a multi-scale-channel Gaussian derivative network on
    training data with sizes 1, 2 and 4, respectively. As can be seen
    from the graphs, the performance is rather similar for all these
    networks over the size range between 1 and 4,  a range for
     which the discretization errors in the discrete implementation can
     be expected to be low (a problem at too fine scales) and the
     influence of boundary effects causing a mismatch between what
     parts of the digits are visible in the testing data compared to a
     training data (a problem at too coarse scales).  The top figure shows the results for the default network
     with 12-16-24-32-64-10 feature channels. The bottom figure shows
     the results for a larger network with 16-24-32-48-64-10 feature
     channels. (Horizontal axis: Scale of testing data.)}
   \label{fig-generalization-all-exp-sizes-1-2-4}
 \end{figure*}

To investigate the properties of a multi-scale-channel architecture experimentally, we
created a multi-scale-channel network with 8 scale channels with their initial
scale values $\sigma_0$ between
$1/\sqrt{2}$ and 8 and a scale ratio of $\sqrt{2}$ between adjacent
scale channels.
For each channel, we used a Gaussian derivative network of similar
architecture as the single-scale-channel network, with 12-14-16-20-64 channels in the
intermediate layers and 10 output channels, and with a relative scale
ratio $r = 1.25$ (\ref{eq-r-geom-distr-sigma}) between adjacent layers, implying that the maximum
value of $\sigma$ in each channel is $\sigma_0 \times 1.25^5 \approx
3.1~\sigma_0$~pixels.

Importantly, the parameters $C_0^{k,c_{out},c_{in}}$, $C_x^{k,c_{out},c_{in}}$, $C_y^{k,c_{out},c_{in}}$, 
$C_{xx}^{k,c_{out},c_{in}}$, $C_{xy}^{k,c_{out},c_{in}}$ and $C_{yy}^{k,c_{out},c_{in}}$ 
in (\ref{eq-lin-comb-2-jet-with indices}) are shared between the scale
channels, implying that the scale channels together are truly scale
covariant, because of the parameterization of the receptive fields in
terms of scale-normalized Gaussian derivatives.  The batch
normalization stage is also shared between the scale channels.
The output from max pooling over the scale channels is furthermore
truly scale invariant, if we assume an infinite number of scale
channels, so that scale boundary effects can be disregarded.

Figure~\ref{fig-generalization-exp-sizes-1-2-4} shows the result of an
experiment to investigate the ability of such a multi-scale-channel
network to generalize to testing at scales not present in the
training data.

For the experiment shown in the top figure, we have used 50~000 training
images for training size 1, and 10~000 testing images for each one of the 19
testing sizes between 1/2 and 8 with a relative size ratio of $\sqrt[4]{2}$
between adjacent testing sizes. The red curve shows the generalization
performance for a single-scale-channel network with $\sigma_0 = 1$,
whereas the blue curve shows the generalization performance for the
multi-scale-channel network with 8 scale channels between $1/\sqrt{2}$ and 8.

As can be seen from the graphs, the generalization performance is
very good for the multi-scale-channel network, for sizes in roughly
in the range $1/\sqrt{2}$ and $4\sqrt{2}$. For smaller testing sizes near 1/2,
there are discretization problems due to sampling artefacts and too
fine scale levels relative to the grid spacing in the image, implying
that the transformation properties under scaling transformations of
the underlying Gaussian derivatives are not well approximated in the
discrete implementation. For larger testing sizes near 8, there are problems
due to boundary effects and that the entire digit is not visible in
the testing stage, implying a mismatch between the training data and
the testing data. Otherwise, the generalization performance is very
good over the size range between 1 and~4.

For the single-scale-channel network, the generalization performance to scales
far from the scales in the training data is on the other hand very
poor.

In the middle figure, we show the result of a similar experiment for
training images of size 2 and with the initial scale level $\sigma_0 = 2$
for the single-scale-channel network.
The bottom figure shows the result of a similar experiment performed with
training images of size 4 and with the initial scale level $\sigma_0 = 4$
for the single-scale-channel network.

\begin{figure*}[hbtp]
\begin{center}
   \begin{tabular}{c}
      {\em Selected scale levels: Larger network trained at size 1} \\
       \includegraphics[width=0.70\textwidth]{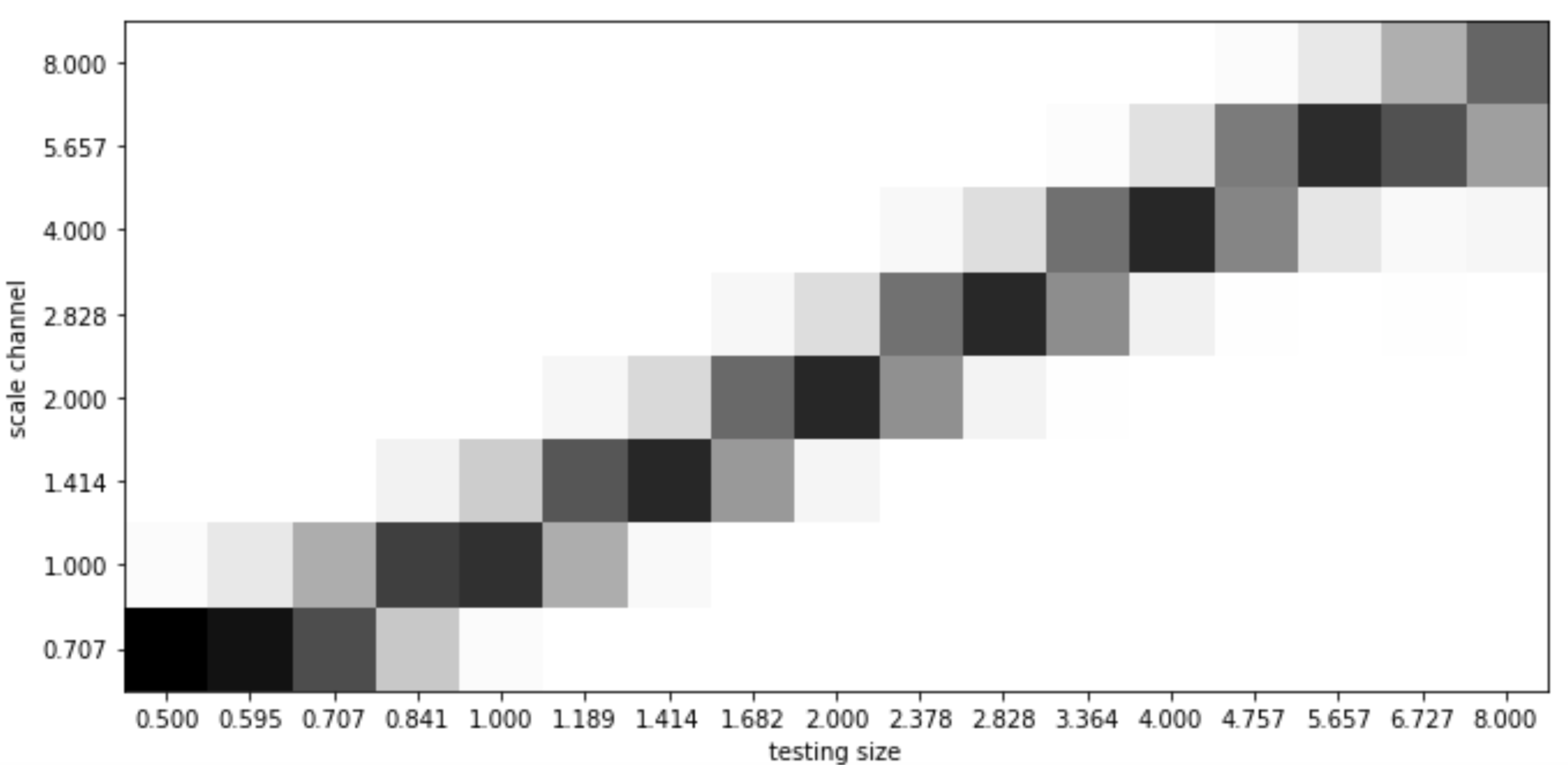}\\
     \\
      {\em Selected scale levels: Larger network trained at size 2} \\
       \includegraphics[width=0.70\textwidth]{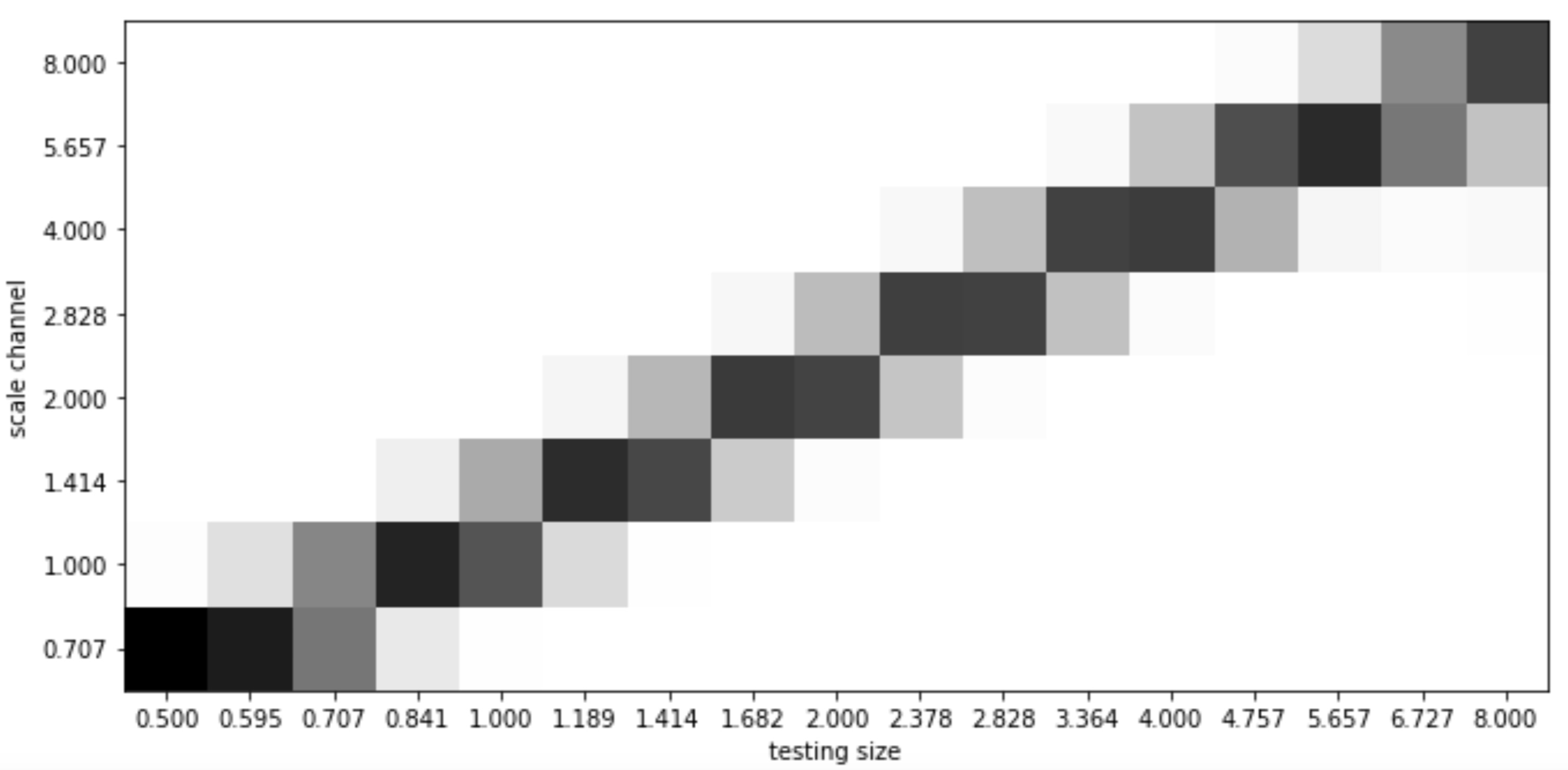}\\
      \\
      {\em Selected scale levels: Larger network trained at size 4} \\
       \includegraphics[width=0.70\textwidth]{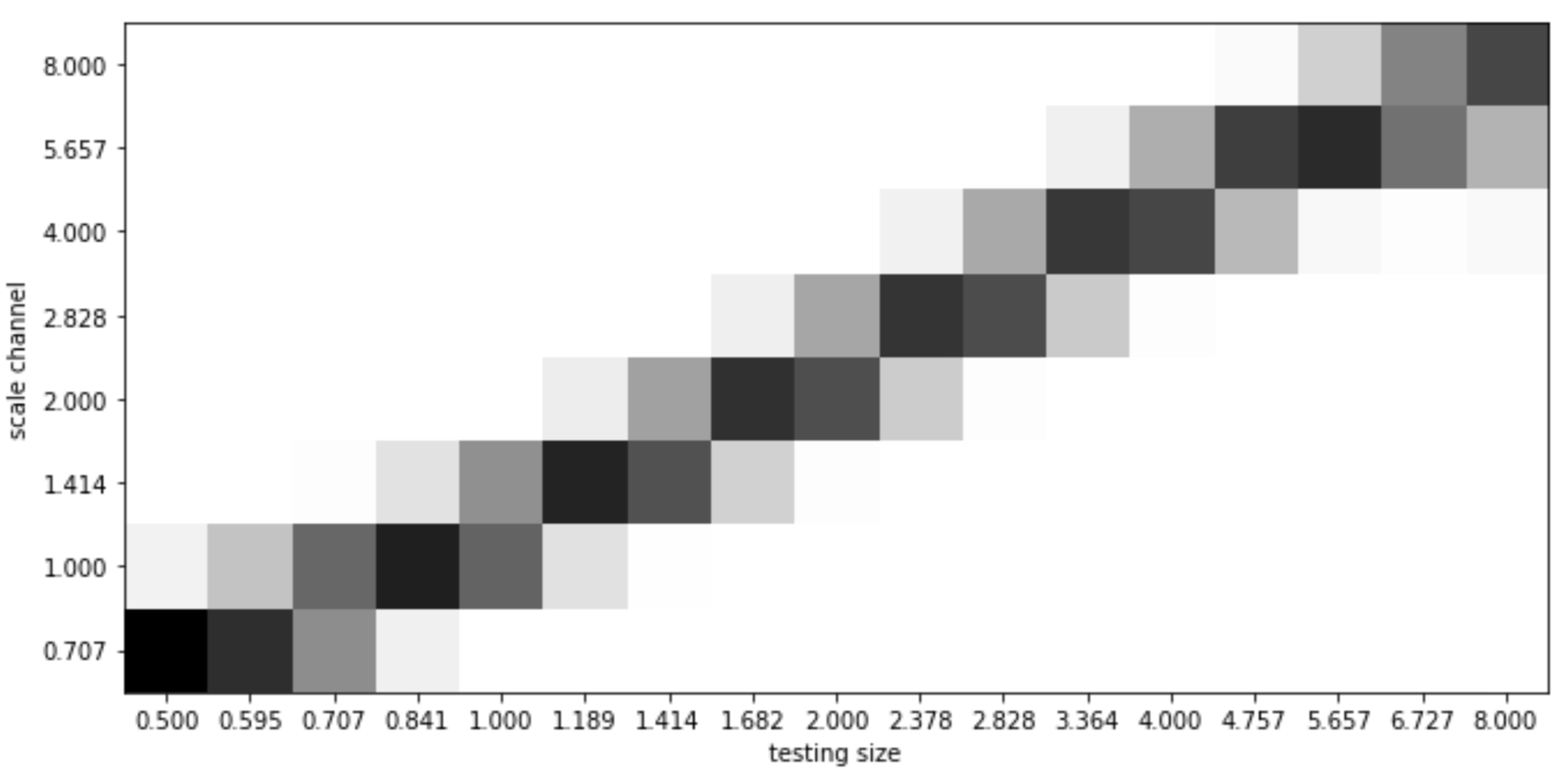}\\
    \end{tabular}
  \end{center}
  \caption{Visualization of the scale channels that are selected in the
    max pooling stage, when training the larger network for each one
    of the sizes 1, 2 and 4. For each testing size, shown on the
    horizontal axis, the vertical axis displays a histogram of the
    scale levels at which the maximum over the scale channels is
    assumed for the different samples in the testing set, with the
    lowest scale at the bottom and the highest scale at the top. As can be
    seen from the figure, there is a general tendency of the
    composed classification scheme to select coarser scale levels with
  increasing size of the image structures, in agreement with the
  conceptual similarity to classical methods for scale selection based
on detecting local extrema over scales of scale-normalized derivatives, the
difference being that here only the global maximum over scale is used,
as opposed to the detection of multiple local extrema over scale in
classical scale selection methods.}
   \label{fig-sc-sel-MNISTLargeScale}
\end{figure*}

Figure~\ref{fig-generalization-all-exp-sizes-1-2-4} shows a joint
visualization of the generalization performance for all these
experiments, where we have also zoomed in on the top performance
values in  the range 98-99~\%.
In addition to the results from the default network with
12-16-24-32-64-10 feature channels, we do also show results obtained
from a larger network with 16-24-32-48-64-10 feature channels, which has
more degrees of freedom in the training stage (a total number of $6 \times (16 + 16 \times 24 + 24 \times 32 + 32 \times 48 + 48
\times 64 + 64 \times 10) = 38~496$ parameters) and leads to higher top
performance and also somewhat better generalization performance. 
As can be seen from the graphs, the performance values for training sizes 1, 2
and 4, respectively, are quite similar for testing data with sizes
in the range between 1 and 4, a size range for which the discretization errors in the discrete implementation can
be expected to be low (a problem at too fine scales) and the
influence of boundary effects causing a mismatch between what
parts of the digits are visible in the testing data compared to the
training data (a problem at too coarse scales).

To conclude, the experiment demonstrates that it is possible to use
the combination of (i)~scale-space features as computational primitives
for a deep learning method with (ii)~the closed-form transformation properties
of the scale-space primitives under scaling transformations to (iii)~make a deep network generalize to
new scale levels not spanned by the training data.

\subsection{Scale selection properties}
\label{sec-sc-sel}

Since the Gaussian derivative network is expressed in terms of
scale-normalized derivatives over multiple scales, and the max-pooling
operation over the scale channels implies detecting maxima over scale,
the resulting approach shares similarities to classical methods for
scale selection based on local extrema over scales of scale-normalized
derivatives \cite{Lin97-IJCV,Lin98-IJCV,Lin21-EncCompVis}. The approach is also closely related to the scale
selection approach in \cite{LooLiTax09-LNCS,LiTaxLoo12-IVC} based on choosing the scales at which a supervised
classifier delivers class labels with the highest posterior.

A limitation of choosing only a single maximum over scales compared
to processing multiple local extrema over scales as in \cite{Lin97-IJCV,Lin98-IJCV,Lin21-EncCompVis}, however,
is that the approach may be sensitive to boundary effects at the scale
boundaries, implying that the scale generalization properties may be
affected depending on how many coarser-scale and/or finer-scale
channels are being processed relative to the scale of the image data.

In Figure~\ref{fig-sc-sel-MNISTLargeScale}, we have visualized the
scale selection properties when applying these multi-scale-channel
Gaussian derivative networks to the MNIST Large Scale dataset.
For each one of the training sizes 1, 2 and 4, we show what scales are
selected as function of the testing size.
Specifically, for each testing size, shown on the horizontal axis,
we display a histogram of the scale channels at which the maximum over
scales is assumed over all the samples in the testing set, with the
finest scale at the bottom and the coarsest scale at the top.

As can be seen from the figure, the network has an overall tendency of
selecting coarser scale levels with increasing size of the image
structures it is applied to. Specifically, the overall tendency is
that the selected scale is proportional to the size of the testing
data, in agreement with the theory for scale selection based on local
extrema over scales of scale-normalized derivatives.

Except for minor quantization variations due to the discrete bins, these
scale selection histograms are very similar for the different training
sizes 1, 2 and 4.
In combination with the previously presented experiments, the results
in this paper
thus demonstrate that it is possible to use scale-space
operations as computational primitives in deep networks, and to use the
transformation properties of such computational primitives under
scaling transformations to perform scale generalization.

\subsection{Properties of multi-scale-channel network training}
\label{sec-mult-sc-train}

While the scale generalization graphs in
Figure~\ref{fig-generalization-all-exp-sizes-1-2-4} and the scale selection
histograms in Figure~\ref{fig-sc-sel-MNISTLargeScale} are very similar for the
different training scales 1, 2 or 4, we can note that
the results are, however, not fully identical.

This can partly be explained from the fact that the training error does
not approach zero (the training accuracy for the larger multi-scale channel
networks trained at a single scale reaches the order of
99.6--99.7~\%), implying that the net effects of the training error on the
properties of the network may be somewhat different when training the
network on different datasets, here for different training sizes.

Additionally, even if the training error would have approached zero,
different types of scale boundary effects and image boundary effects could occur for the
different networks, depending on 
what single size training data the multi-scale channel network is
trained on, implying that the learning algorithm could lead to
sets of filter weights in the networks with somewhat different
properties, because of a lack of full scale covariance in the training
stage, although the architecture of the network is otherwise fully
scale covariant in the continuous case.

The training process is initiated randomly, and during the gradient
descent optimization, the network has to learn to associate large
values of the feature map for the appropriate class for each training sample in some scale channel, while
also having to learn to not associate large values for erroneous
classes for any other training samples in any other scale channel. 

Training of a multi-scale-channel
network could therefore be considered as a harder training problem
than the training of a single-scale-channel network. Experimentally, we have
also observed that the training error is significantly larger for a
multi-scale-channel network than for a single-scale-channel network,
and that the training procedure had not converged fully after the 20
epochs that we used for training the multi-scale-channel networks.

A possible extension of this work, is therefore to perform a deeper
study regarding the task of training multi-scale-channel networks, and
investigate if this training task calls for other training strategies
than used here, specifically considering that the computational
primitives in the layers of the Gaussian derivative networks are also
different from the computational primitives in regular CNNs,
while we have here used a training strategy suitable for regular CNNs.

Notwithstanding these possibilities for improvements in the training scheme, the experiments
in the paper demonstrate that it is possible to use scale-space
operations as computational primitives in deep networks, and to use the
transformation properties of such computational primitives under
scaling transformations to perform scale generalization, which is the
main objective of this study.

\section{Summary and discussion}
\label{sec-summ-disc}

We have presented a hybrid approach between scale-space theory and
deep learning, where the layers in a hierarchical network architecture
are modelled as continuous functions instead of discrete filters,
and are specifically chosen as scale-space operations, here in terms of
linear combinations of scale-normalized Gaussian derivatives.
Experimentally, we have demonstrated that the resulting approach
allows for scale generalization and enables good performance for
classifying image patterns at scales not spanned by the training data.

The work is intended as a proof-of-concept of the idea of using
scale-space features as computational primitives in a deep learning
method, and of using their closed-form transformation properties under
scaling transformations to perform extrapolation or generalization to new scales not
present in the training data. 

Concerning the choice of Gaussian derivatives as computational
primitives in the method, it should, however, be emphasized that the
necessity results that specify the uniqueness of these kernels only state that the first
layer of receptive fields should be constructed in terms of Gaussian
derivatives. Concerning higher layers, further studies should be
performed concerning the possibilities of using other scale-space
features in the higher layers, that may represent the variability of
natural image structures more efficiently, within the generality of
the sufficiency result for scale-covariant continuous hierarchical networks in \cite{Lin20-JMIV}.

\section*{Acknowledgements}

I would like to thank Ylva Jansson for sharing her code for training and
testing networks in PyTorch and for valuable comments on an earlier
version of this manuscript.

I would also like to thank the anonymous reviewers for valuable
comments, which helped to improve the presentation.

\vspace{-4mm}

$\,$ 

{\footnotesize
\bibliographystyle{splncs}
\bibliography{defs,tlmac}}

\end{document}